\documentclass[journal]{IEEEtran}

\usepackage{graphics} 
\usepackage{epsfig} 
\usepackage{times} 
\usepackage{amsmath} 
\usepackage{amssymb}  
\usepackage{bm}
\usepackage{multirow}
\usepackage{graphicx}
\usepackage{float}
\usepackage{subfig}
\usepackage{url}
\usepackage{hyperref}
\hypersetup{hypertex=true,
hidelinks=true,
urlcolor=blue,
colorlinks=true,
linkcolor=blue,
anchorcolor=blue,
citecolor=blue}
\usepackage{flushend}
\usepackage{booktabs} 
\usepackage{bbding}
\usepackage{xcolor} 
\newcommand{\mathcolorbox}[2]{\colorbox{#1}{$\displaystyle #2$}}
\usepackage{soul} 
\definecolor{lightblue}{rgb}{0.812, 0.906, 0.922}
\definecolor{lightgreen}{rgb}{0.651, 0.780, 0.415}
\definecolor{lightorange}{rgb}{0.984, 0.757, 0.678}
\definecolor{FFDF70}{rgb}{1.000, 0.875, 0.439}
\definecolor{F46E49}{rgb}{0.956863, 0.431373, 0.286275}
\definecolor{81D0BB}{rgb}{0.505882, 0.815686, 0.733333}
\definecolor{CFE7EB}{rgb}{0.811765, 0.905882, 0.921569}
\definecolor{FBC1AD}{rgb}{0.984314, 0.756863, 0.678431}
\definecolor{7DABCF}{rgb}{0.490196, 0.670588, 0.811765}
\definecolor{377E22}{rgb}{0.216, 0.494, 0.133}
\definecolor{E76D7E}{rgb}{0.909804, 0.427451, 0.494118}
\definecolor{F2B1BB}{rgb}{0.949, 0.694, 0.733}
\definecolor{B6DDBB}{rgb}{0.714, 0.867, 0.733}
\definecolor{45AAB4}{rgb}{0.270, 0.666, 0.706}
\definecolor{377E22}{rgb}{0.215, 0.494, 0.133}
\usepackage[table]{xcolor} 
\newcommand{\best}[1]{\cellcolor{green!15}#1}
\newcommand{\second}[1]{\cellcolor{blue!10}#1}
\newcommand{\cmark}{\checkmark}

\usepackage{tcolorbox}

\newcommand{\pill}[3]{%
  \tcbox[
    on line,
    colback=#1,
    colframe=#1,
    boxrule=0pt,
    arc=3pt,          
    top=1pt,
    bottom=1pt,
    left=3pt,
    right=3pt,
    boxsep=0pt,
    fontupper=\sffamily\small
  ]{\textcolor{#2}{#3}}%
}

\newcommand{\bestt}[1]{\pill{green!20}{black}{#1}}
\newcommand{\secondt}[1]{\pill{blue!15}{black}{#1}}

\usepackage{tikz}     

\newcommand{\trajectorySegment}[2]{
    {\begin{tikzpicture}[scale=0.2, baseline=-0.6ex]
        \foreach \i in {0,...,3} {
            \ifnum\i<#1
                \filldraw[fill=white] (\i,0) rectangle ++(1,1);
            \else\ifnum\i<#2
                \filldraw[fill=gray] (\i,0) rectangle ++(1,1);
            \else
                \filldraw[fill=white] (\i,0) rectangle ++(1,1);
            \fi\fi
        }
    \end{tikzpicture}}
}

\newcommand{\trajectorySegmentExplained}[2]{%
{%
\begin{tikzpicture}[scale=0.2, baseline=-6.5ex]
    \foreach \i in {0,...,3} {
        \ifnum\i<#1
            \filldraw[fill=white] (\i,0) rectangle ++(1,1);
        \else\ifnum\i<#2
            \filldraw[fill=gray] (\i,0) rectangle ++(1,1);
        \else
            \filldraw[fill=white] (\i,0) rectangle ++(1,1);
        \fi\fi
    }

    \draw[thick] (0.5,0.5) -- (0.5,-6.0) -- (6.0,-6.0);
    \node[right, inner sep=1pt] at (6.0,-6.0) {\(\scriptscriptstyle [-T+1,-1]\)};

    \draw[thick] (1.5,0.5) -- (1.5,-4.4) -- (6.0,-4.4);
    \node[right, inner sep=1pt] at (6.0,-4.4) {\(\scriptscriptstyle 0\)};

    \draw[thick] (2.5,0.5) -- (2.5,-2.8) -- (6.0,-2.8);
    \node[right, inner sep=1pt] at (6.0,-2.8) {\(\scriptscriptstyle [1,H-1]\)};

    \draw[thick] (3.5,0.5) -- (3.5,-1.2) -- (6.0,-1.2);
    \node[right, inner sep=1pt] at (6.0,-1.2) {\(\scriptscriptstyle H\)};
\end{tikzpicture}%
}%
}

\title{ 
Robot Trajectron V3: A Probabilistic Shared Control Framework for $SE(3)$ Manipulation
}

\author{Pinhao Song$^{1,3}$, Zhongxi Li$^{2,3}$, Ze Fu$^{1,3}$, Federico Ulloa Rios$^{1,3}$, Renaud Detry$^{1,2,3}$
\thanks{Supported by Interne Fondsen KU Leuven/Internal Funds KU Leuven. Partially supported by Flanders Make (strategic research centre for the manufacturing industry).}
\thanks{$^{1}$KU Leuven, Dept. Mechanical Engineering, Research unit \emph{Robotics, Automation and Mechatronics}, B-3000 Leuven, Belgium.
        {\tt\small Email: firstname.lastname@kuleuven.be}}
\thanks{$^{2}$KU Leuven, Dept. Electrical Engineering, Research unit \emph{Processing Speech and Images}, B-3000 Leuven, Belgium.
}
\thanks{$^{3}$Flanders Make@KU Leuven.}
}

\begin{document}

\maketitle
\thispagestyle{empty}
\pagestyle{empty}
\begin{abstract}

We aim to address the challenge of teleoperating robotic arms for high-degree-of-freedom (high-DoF) manipulation tasks, which is cognitively demanding and error-prone, particularly when relying on low-bandwidth interfaces. We propose Robot Trajectron V3 (RT-V3), a probabilistic shared control framework designed for $SE(3)$ grasping tasks. RT-V3 formulates shared control as Bayesian inference by learning a prior over user intent and combining it with real-time user commands to estimate the posterior intent distribution. The prior models user intent as a distribution over future trajectories conditioned on past robot dynamics and visual scene context. The intent prior is parameterized by a transformer-based conditional generative model that reasons over point clouds and candidate grasp poses, together with a factorized translation-rotation representation that improves learning efficiency in high-dimensional action spaces. During execution, RT-V3 continuously estimates the posterior distribution over future trajectories by combining the learned intent prior with a user-command likelihood derived from the observed control input, enabling continuous intent refinement and shared assistance. Comprehensive experiments demonstrate that RT-V3 achieves high accuracy in trajectory prediction and competitive performance in reactive planning. Furthermore, real-world user studies indicate that RT-V3 significantly outperforms baseline methods in terms of success rate and efficiency, while substantially reducing the user's physical and mental workload. 
Our code is available at \url{https://mousecpn.github.io/RTV3_page/}.


\end{abstract}

\section{Introduction}
Teleoperation of robotic arms has made significant progress in recent years, improving both real-time performance and usability. Modern teleoperation systems often rely on high-bandwidth interfaces such as shadow arms \cite{fu2024mobile}, virtual reality devices \cite{cheng2024open}, or motion capture systems \cite{gan2024telemotion}, enabling users to control robots with high precision. Such systems provide an important pathway for assistive manipulation, allowing individuals to interact with and manipulate objects through robotic devices. However, these interfaces are often inaccessible to users with motor impairments. 
Instead, such users typically rely on low-bandwidth and noisy input devices, such as chin joysticks \cite{rulik2022control} or head joysticks \cite{rofer2009controlling}. Emerging neural interfaces \cite{hochberg2012reach,saussus2026stabilization} further offer a promising alternative for users with severe motor impairments, but they remain limited by low-bandwidth and noisy control signals.
Controlling high-degree-of-freedom (high DoF) robotic manipulators with these interfaces is challenging. Performing manipulation tasks such as grasping objects from specific orientations often requires numerous commands, making teleoperation slow, cognitively demanding, and prone to error. Shared control addresses this challenge by combining user input with contextual information to infer user intent and assist task execution. By integrating human commands with environmental perception, shared control systems can reduce user effort while preserving user authority, enabling more efficient human–robot collaboration.

Shared autonomy methods have demonstrated promising results in planar navigation \cite{psc, lei2022intention} and simplified manipulation tasks \cite{gottardi2022shared}. However, shared control for robotic grasping in realistic environments remains challenging. Two key factors contribute to this difficulty.

\noindent \textbf{The complexity of the environment.} Robotic manipulation often occurs in cluttered environments containing multiple objects with diverse affordances. Assistive algorithms must reason about which objects are actionable and infer the user's intended target among several candidates. Prior work often simplifies this setting by considering only one or two sparsely placed objects \cite{policyblending, wang2018continuous, oh2020natural}, which does not reflect real-world scenarios. Furthermore, a single object may support multiple valid grasp poses depending on the intended task, whereas many existing approaches assume a single grasp configuration per object \cite{gottardi2022shared, lee2025brain, xu2020shared, RT}. Such simplifications limit the ability of shared control systems to accurately capture user intent.

\noindent \textbf{Low-bandwidth user input on a high DoF system.} Shared control systems must bridge the mismatch between low-bandwidth user input and high-DoF robotic manipulators. Manipulation tasks typically require controlling a full $SE(3)$ end-effector pose, while assistive interfaces provide only a small number of control channels. As a result, directly commanding all degrees of freedom is often inefficient and exhausting for users. Existing work commonly reduces the problem by restricting the manipulation strategy, such as enforcing top-down grasps \cite{gottardi2022shared, lee2025brain} or side grasps \cite{policyblending, oh2020natural}. While these simplifications reduce control complexity, they limit the diversity of achievable grasps and restrict the range of tasks that can be performed. 

To address these challenges, we propose Robot Trajectron V3 (RT-V3), a probabilistic shared control framework for multi-object, multi-affordance $SE(3)$ grasping assistance under low-bandwidth user interfaces. RT-V3 formulates shared control within a Bayesian inference framework:
\begin{equation}
    p(i|u,c) \propto p(u|i)p(i|c), \label{eq: psc}
\end{equation}
where $i$ denotes the user’s behavioral intent, represented as a future trajectory, $u$ denotes the current user command, and $c$ denotes the environmental context. The prior $p(i|c)$ models likely user behaviors given the scene context, while the likelihood $p(u|i)$ captures the uncertainty of the user interface. Shared control is achieved by estimating the posterior intent distribution and assisting the user accordingly. The context representation $c$ includes both geometric and affordance information. We represent scene geometry using point clouds and encode candidate grasp configurations as grasp pose clouds generated by a grasp planner. Based on this representation, the prior model predicts plausible future trajectories conditioned on the scene context. The proposed RT-V3 builds upon the Robot Trajectron V2 \cite{rtv2} navigation shared autonomy framework, extending it to the domain of robotic manipulation.

To effectively model this prior, we introduce a transformer-based architecture that encodes the unstructured $SE(3)$ scene representation. The model jointly reasons over point clouds, grasp candidates, and robot dynamics through cross-attention mechanisms, enabling the system to identify relevant objects and affordances for intent inference. In addition, we factorize the action distribution into a translational component and a rotational component conditioned on translation. This factorization reduces the complexity of density estimation and enhances stability during closed-loop execution.

Finally, we introduce an asynchronous shared control mechanism that leverages periods without user input to update intent inference and shared policies. By continuously refining the posterior estimate of user intent, the system can proactively assist the user and reduce the number of required commands, alleviating fatigue associated with low-bandwidth interfaces.

The contributions of this work are threefold:
\begin{itemize}
    \item A probabilistic shared control framework for multi-object, multi-affordance $SE(3)$ manipulation, extending Robot Trajectron from navigation to robotic grasping.
    \item A transformer-based architecture for encoding unstructured $SE(3)$ scene context, enabling reasoning over point clouds and grasp affordances for intent prediction.
    \item A factorized action representation and asynchronous shared control mechanism, which improves sample efficiency, closed-loop stability, and user interaction efficiency.
\end{itemize}

We evaluate RT-V3 through trajectory prediction benchmarks, closed-loop planning experiments, and human user studies using joystick teleoperation. The results demonstrate that RT-V3 achieves accurate intent prediction and significantly improves shared control performance in terms of efficiency and user agreement.

\section{Related Work}
\subsection{Shared Control}
Shared control involves the cooperative determination of a policy by both an autonomous agent and the user, with the goal of improving performance and safety in robot manipulation. This approach is widely utilized in various applications, including assistive driving \cite{lu2019model}, wheelchair control \cite{vanhooydonck2003shared}, and robotics manipulation \cite{song2026assistron}. Shared control for high-DoF robotic arms has been widely explored. Gottardi et al.~\cite{gottardi2022shared} proposed an improved artificial potential field method to guide the robot toward likely goals while avoiding obstacles. Robot Trajectron \cite{RT} predicts the future end-effector trajectory from recent motion dynamics, enabling proactive support for reaching tasks. Oh et al. \cite{oh2020natural} regard shared control as a Q-value maximization problem with a divergence constraint between the robot and the shared policy, and solve the problem by natural gradients. However, these methods are limited to 2D or 3D translation space, and providing full $SE(3)$ assistance is still challenging. Hindsight Optimization (HO) \cite{javdani2018shared} provides an inverse reinforcement learning (IRL) perspective to solve the intent estimation problem, which has inspired numerous shared control approaches and shown promising performance even in complex, cluttered environments \cite{muelling2017autonomy, gottardi2022shared, yow2023shared, fu2025tasc}. However, hindsight optimization can only predict the probability for each potential target. Therefore, planning and policy blending to transform the target probabilities into real-time action guidance are still needed. Besides, HO's predictions largely rely on Euclidean distance between the current position and the potential target poses without a consideration of motion dynamics \cite{RT} and collision avoidance, leading to inaccurate predictions sometimes.

Recently proposed Robot Trajectron V2 (RT-V2) \cite{rtv2} provides a probabilistic shared control framework. Instead of predicting the final target, RT-V2 seeks a behavioral modeling perspective to solve the shared control problem. It models the user's behavior patterns, predicts collision-free future actions, and blends the user's command via posterior estimation. However, RT-V2 is limited to 2D translation space. In this work, we propose Robot Trajectron V3, which lifts RT-V2 from 2D translation space to $SE(3)$. Compared to RT-V2, which uses CNNs to encode the context in occupancy map format, RT-V3 leverages a transformer-based encoder to encode point clouds and grasp poses. Besides, RT-V3 factorizes the action distribution into a translation distribution and a rotation distribution conditioned on the translation.

\subsection{Imitation Learning}

Imitation Learning (IL), also referred to as Learning from Demonstration (LfD) or Behavior Cloning (BC), aims to train policies that mimic expert behavior. IL has been widely applied across robotics domains due to its efficiency in leveraging demonstrations for data collection and policy training \cite{ILsurvey, ILsurveyDriving, rtx}. Despite its appeal, several challenges remain in adapting IL to complex robotic manipulation tasks.

\textbf{Modeling multimodal action distributions.} Robotic tasks often admit multiple valid strategies, making the underlying action distribution inherently multimodal. Generative models are well-suited to capture such distributions, and recent work has integrated them into IL. Conditional variational autoencoders (CVAEs) \cite{act}, energy-based models \cite{ibc}, and diffusion models \cite{chi2023diffusion} have all been proposed to represent multimodal expert behaviors. These approaches have demonstrated strong performance on standard robotic benchmarks such as RLBench \cite{james2020rlbench}, LIBERO \cite{liu2023libero}, and CALVIN \cite{mees2022calvin}.

\textbf{Context representation.} A second key design choice in IL is how the environment is encoded. RGB cameras are widely used due to their low cost and accessibility, with visual features commonly extracted using CNNs. More recently, large-scale pretrained transformers such as DINOv2 \cite{oquab2023dinov2} have emerged as powerful visual encoders, offering compatibility with transformer-based policies \cite{kim2024openvla, act}. Depth cameras provide complementary 3D information, enabling point cloud representations that can be processed by architectures such as PointNet \cite{qi2017pointnet} and Point Transformer \cite{zhao2021point, wu2022point, wu2024point}. Alternatively, volumetric encodings such as truncated signed distance functions can be learned with 3D CNNs \cite{wangequivariant}. Selecting the appropriate context modality and representation remains an open challenge, as it strongly influences generalization and sample efficiency in IL.

\textbf{Action representation.} Finally, IL performance depends critically on how the robot's action space is represented. Options vary along several dimensions: the space of execution (e.g., $SE(2)$ vs. $SE(3)$); the type of control (position \cite{chi2023diffusion} vs. velocity \cite{sakaino2021bilateral}); continuity (continuous \cite{ke20243d, hu2025mini} vs. discrete \cite{peract}); and the reference frame (absolute \cite{chi2023diffusion} vs. relative \cite{wangequivariant}). For $SE(3)$ actions, rotation can be parameterized in multiple ways, such as Euler angles \cite{kim2021transformer}, quaternions \cite{song2024implicit}, or 6D rotation vectors \cite{ke20243d}. The choice of action representation not only affects downstream performance but also constrains model architecture and learning dynamics.

Inspired by the IL works mentioned above, RT-V3 adopts a CVAE framework to capture the multimodality of user behavior. To handle unstructured $SE(3)$ contextual data, including grasp poses and point clouds, we employ a transformer-based encoder that integrates spatial and temporal information. Finally, to remain compatible with user twist commands and the requirements of shared control, RT-V3 defines its action space in $\mathfrak{se}(3)$ velocity, allowing the policy to directly predict translational and rotational velocities in a geometrically consistent manner.

\begin{figure*}[!t]
  \centering
  \includegraphics[width=0.8\linewidth]{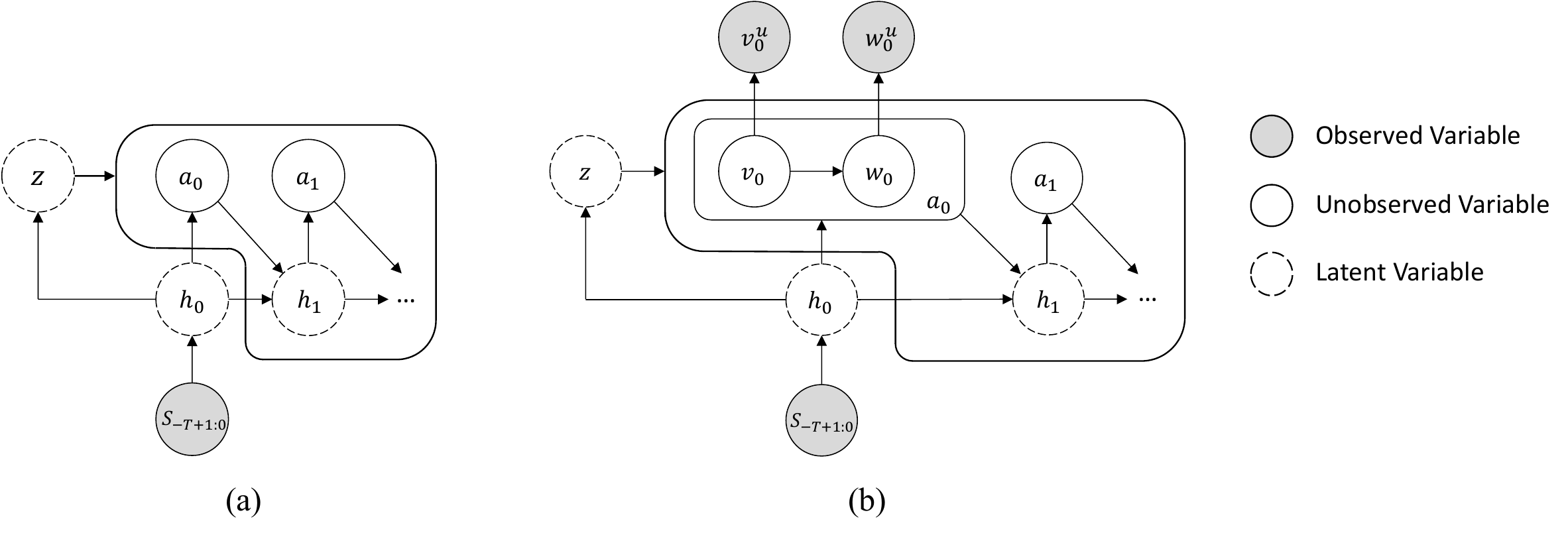}
  \caption{
  Graphical models of the probabilistic decision-making process in the RT-V3 framework. Trajectories are represented using relative timesteps, where $t=0$ denotes the current timestep. (a) The prior model: This model captures the user's inherent behavioral patterns. The past state trajectory and unstructured context ($\bm{s}_{-T+1:0}$) are encoded into an initial recurrent state ($\bm{h}_0$). To account for the multi-modality of manipulation intents, a discrete latent variable ($\bm{z}$) is introduced, which governs the generation of future hidden states ($\bm{h}_t$) and corresponding actions ($\bm{a}_t$). (b) The posterior model: During closed-loop shared control at $t=0$, the system observes a noisy user command $\bm{a}_0^u$, which consists of translational ($\bm{v}_0^u$) and rotational ($\bm{w}_0^u$) velocities. The intended action $\bm{a}_0$ is explicitly factorized into a translational component ($\bm{v}_0$) and a rotational component ($\bm{w}_0$) conditioned on translation. By fusing the uncertain user command with the learned behavioral prior via Bayesian posterior estimation, RT-V3 continuously infers the most probable intended action. 
  }
  \label{fig: graphical model}
\end{figure*}

\section{Methodology Overview}
In this work, we study shared control for robotic grasping, where a human user operates a robot to grasp an object in the workspace with assistance from an autonomous controller. The controller does not know in advance which object the user intends to grasp or how the grasp should be executed. Instead, it must infer the user’s intent from past input commands, the robot’s motion history, and contextual sensory information. 

We leverage off-the-shelf grasp planners \cite{song2024implicit, song2025equivariant}, which provide a dense set of candidate grasp poses $\mathcal{G}$ for objects in the scene. \textbf{We assume that this dense set of candidate grasp poses includes the grasp poses that the user desires to execute}. This assumption allows us to formulate shared-control grasping as a goal-pose reaching task: the grasping workspace can be abstracted as a set of $M$ candidate destinations $\mathcal{G} \in \mathbb{R}^{M \times 3}$ and a point cloud with $N$ points $\mathcal{P}  \in \mathbb{R}^{N \times 6}$. Operating within this defined workspace, the user and the robot collaborate through a shared-control paradigm. 
To explicitly delineate these collaboration dynamics—aligning with the taxonomy established by Losey et al. \cite{losey2018review}—we distinguish between the user's unobservable goal, the measurable interface signals, and the system's internal prediction. During operation, the user possesses an unobserved \textit{ultimate goal} (one of the candidate destinations) that they desire to reach. They express this intent by issuing continuous \textit{user commands} ($\bm{u}$), which are instantaneous 6D translational and rotational velocity signals transmitted through a physical, inherently noisy interface (e.g., joystick, ...). The assistive controller, RT-V3, gathers these noisy commands along with the environmental context to iteratively compute an \textit{estimated intent} ($\bm{i}$). Rather than just predicting a discrete target, this estimated intent is modeled as an \emph{intended future trajectory}—a sequence of desired, noise-free velocity actions that bridge the robot's current state to the ultimate goal. By separating the underlying signal from the interface noise, the system ultimately issues goal-directed robot velocity commands that realize the perceived user intent.


Our method approaches shared control from a Bayesian perspective. The methodology, detailed in Sections \ref{sec. psc} through \ref{posterior}, is organized as follows:

Section~\ref{sec. psc}: We detail the formulation of the probabilistic shared control framework described in Eq.~\ref{eq: psc}. In this framework, shared control is achieved by posterior estimation $p(i|u,c)$, where the prior model for posterior estimation $p(i|c)$ is obtained by training a model in a supervised way to predict how the onset of a robot motion is likely to continue. The model's input is a ``past'' trajectory and contextual cues, and its output is the robot's ``future'' trajectory.


Section~\ref{sec. arch}: We explain the parametrization of the proposed prior model. Specifically, the prior model is based on conditional variational auto-encoder frameworks, in which we use RNNs to encode the dynamic information and use a transformer-based architecture to encode the unstructured context. Besides, in the decoding of the future trajectories, we propose a translation-conditioned rotation architecture, which simplifies density estimation and improves stability during closed-loop execution. 

Section~\ref{posterior}: This section will explain how to achieve posterior estimation $p(i|u,c)$ of the future trajectories given the current \emph{user input} in our parametrization setting mentioned in Sec.~\ref{sec. arch}. Specifically, we adopt an agentic perspective, where the user provides a noisy command $u$ at each timestep. We model the likelihood $p(u|i)$ and combine it with the prior $p(i|c)$ (the intent estimator) to obtain the posterior $p(i|u,c)$.



\section{Probabilistic Shared Control} \label{sec. psc}
In this section, we detail the theoretical formulation of the probabilistic shared control framework of RT-V3. Note that this section focuses purely on the mathematical objectives—specifically, defining the prior and the posterior distribution. The specific neural network parameterizations used to learn this prior are deferred to Sec.~\ref{sec. arch}, and the computation of the posterior during real-time teleoperation is detailed in Section Sec.~\ref{posterior}.

The prior model of RT-V3 is designed to encode the user's inherent motion behaviors. Let us denote with the variable $\bm{s}_t$ the state at $t$, incorporating the current dynamics and contextual information, and with $\bm{a}_t$ the velocity action at $t$. Using $t=0$ to represent the \emph{current} timestep, we write the intended trajectory as $\bm{\zeta} = (\bm{s}_{-T+1:H},\bm{a}_{-T+1:H-1})$, in which $(\bm{s}_{-T+1:0},\bm{a}_{-T+1:-1})$ is the \emph{past} trajectory and $(\bm{s}_{1:H},\bm{a}_{0:H-1})$ is the \emph{future} trajectory. The intended trajectories can be sampled from the interaction between the user and the environment with failure samples filtered out. The prior model is obtained via the following goal: 
\begin{equation}
\mathop{\textnormal{max}}~\mathbb{E}_{\bm{\zeta}} [\textnormal{log}~p(\bm{s}_{1:H}, \bm{a}_{0:H-1}|\bm{s}_{-T+1:0} )].  \label{eq: mle}
\end{equation}
The goal of Eq.~\ref{eq: mle} is to maximize the log-likelihood of the intended trajectories. To decompose Eq.~\ref{eq: mle}, we introduce a latent state $\bm{h}_t$, which encodes a memory of past states $\bm{s}_{\leq t}$, then we can simplify $p(\bm{s}_{1:H}, \bm{a}_{0:H-1}|\bm{s}_{-T+1:0})$ as $p(\bm{h}_{0:H}, \bm{a}_{0:H-1}|\bm{s}_{-T+1:0})$. We assume that the latent state obeys a Markovian property: the latent state $\bm{h}_t$ only depends on the previous latent state $\bm{h}_{t-1}$ and the previous action $\bm{a}_{t-1}$. With this assumption, we can decompose $p(\bm{h}_{0:H}, \bm{a}_{0:H-1}|\bm{s}_{-T+1:0})$ as:
\begin{equation}
\begin{aligned}
    & p(\bm{h}_{0:H}, \bm{a}_{0:H-1}|\bm{s}_{-T+1:0})\\ 
    & =   p(\bm{h}_0|\bm{s}_{-T+1:0}) \prod_{t=0}^{H-1}  {p(\bm{h}_{t+1}|\bm{h}_{t}, \bm{a}_{t})} {p(\bm{a}_{t}|\bm{h}_{t})},
    \label{eq: decomposed mle2}
\end{aligned}
\end{equation}
where ${p(\bm{h}_{t+1}|\bm{h}_{t}, \bm{a}_{t})}$ and ${p(\bm{a}_{t}|\bm{h}_{t})}$ are the state transition distribution and the action distribution, respectively.

To model the high multi-modality of the user's intent, we mimic the CVAE framework \cite{cvae,RT} and introduce a discrete latent variable $\bm{z}$, which obeys a categorical distribution, to facilitate the encoding of a low-dimensional, multi-modal representation, as:
\begin{equation}
\begin{aligned}
    & p(\bm{h}_{0:H}, \bm{a}_{0:H-1}|\bm{s}_{-T+1:0})\\
    & = p(\bm{h}_{0}|\bm{s}_{-T+1:0}) p(\bm{h}_{1:H},\bm{a}_{0:H-1}|\bm{h}_{0})\\
    & =   \sum_{\bm{z}} p(\bm{h}_{0}|\bm{s}_{-T+1:0}) p(\bm{h}_{1:H}, \bm{a}_{0:H-1}|\bm{h}_{0},\bm{z}) p(\bm{z}|\bm{h}_{0})\\
     & =    p(\bm{h}_{0}|\bm{s}_{-T+1:0}) \sum_{\bm{z}} p(\bm{z}|\bm{h}_{0})\\
     &~~~~~~~~~~~~~~\prod_{t=0}^{H-1}  {p(\bm{h}_{t+1}|\bm{h}_{t}, \bm{a}_{t}, \bm{z})} {p(\bm{a}_{t}|\bm{h}_{t}, \bm{z})}.
    \label{eq: decomposed mle with latent}
\end{aligned}
\end{equation}
The graphical model of Eq.~\ref{eq: decomposed mle with latent} can be represented in Fig.~\ref{fig: graphical model}~(a). This latent variable $\bm{z}$ denotes the \emph{maneuver class} reflecting the probable maneuver directions for the user to execute \cite{rtv2}. Compared to Eq.~\ref{eq: decomposed mle2}, Eq.~\ref{eq: decomposed mle with latent} introduce a \emph{maneuver class prior} $p(\bm{z}|\bm{h}_{0})$, and rewrite the ${p(\bm{h}_{t+1}|\bm{h}_{t}, \bm{a}_{t})}$ and  ${p(\bm{a}_{t}|\bm{h}_{t})}$ as ${p(\bm{h}_{t+1}|\bm{h}_{t}, \bm{a}_{t}, \bm{z})}$ and  ${p(\bm{a}_{t}|\bm{h}_{t}, \bm{z})}$, respectively.

To train the model, we maximize the likelihood in Eq.~\ref{eq: decomposed mle with latent} by minimizing, per CVAE practice \cite{beta-vae,cvae}, the $\beta$-weighted evidence-based lower bound (ELBO) loss:
\begin{equation}
\begin{aligned}
     & \mathcal{L}_{\text{ELBO}} \\
     & =  ~ -\mathbb{E}_{\bm{z} \sim p(\bm{z}|\bm{h}_{H})}[\textnormal{log}~p(\bm{h}_{0:H}, \bm{a}_{0:H-1}|\bm{s}_{-T+1:0}, \bm{z})] \\
    &~~ + \beta D_{KL}(q(\bm{z}|\bm{h}_{H})||p(\bm{z}|\bm{h}_{0})) \\
    & =   - \mathbb{E}_{\bm{z} \sim p(\bm{z}|\bm{h}_{H})}[\textnormal{log}~p(\bm{h}_{0}|\bm{s}_{-T+1:0})+ \\ 
    & ~~~~~~~~~~~~~~~\sum_{t=0}^{H-1} \textnormal{log}~ p(\bm{h}_{t+1}|\bm{h}_{t}, \bm{a}_{t},\bm{z}) p(\bm{a}_{t}|\bm{h}_{t},\bm{z})] \\
    &~~ + \beta D_{KL}(q(\bm{z}|\bm{h}_{H})||p(\bm{z}|\bm{h}_{0})).
    \label{eq: ELBO}
\end{aligned}
\end{equation}
where $\bm{h}_{H}$ is the latent state concluding both the past and future trajectories, and we propose to use a \emph{maneuver class posterior} $q(\bm{z}|\bm{h}_{H})$ to guide the training of \emph{maneuver class prior} $p(\bm{z}|\bm{h}_{0})$ by minimizing their KL divergence. By training with the loss function described in Eq.~\ref{eq: ELBO}, RT-V3 models the decision-making distribution $p(\bm{h}_{0:H},\bm{a}_{0:H-1}|\bm{s}_{-T+1:0})$, capturing the behavioral pattern of the intention policy, the ideal policy envisioned by the user.

At the current timestep, we observe the user command $\bm{a}^{\text{u}}_{0}$, which is generated through an uncertain interface. To achieve a more accurate estimation of the intended action sequence $\{\bm{a}_t\}_{t=0}^{H-1}$, we incorporate the observed user command $\bm{a}^{\text{u}}_{0}$ and infer the posterior decision-making distribution as follows:
\begin{equation}
     p(\bm{h}_{0:H}, \bm{a}_{0:H-1}|\bm{s}_{-T+1:0}, \bm{a}^{\text{u}}_{0}). \label{eq: posterior decision making}·
\end{equation}
Taking the argmax of the posterior estimate in Eq.~\ref{eq: posterior decision making} yields the intended action sequence $\bm{a}_{-T+1:H-1}$, allowing us to execute the initial action $\bm{a}_{0}$. However, computing this posterior directly is intractable without a specific parameterization of the prior and the likelihood. Therefore, we first detail the parameterization of our learned prior in Sec.~\ref{sec. arch}, followed by the tractable computation of the posterior distribution in Sec.~\ref{posterior}.

\begin{figure*}[!t]
  \centering
  \includegraphics[width=\linewidth]{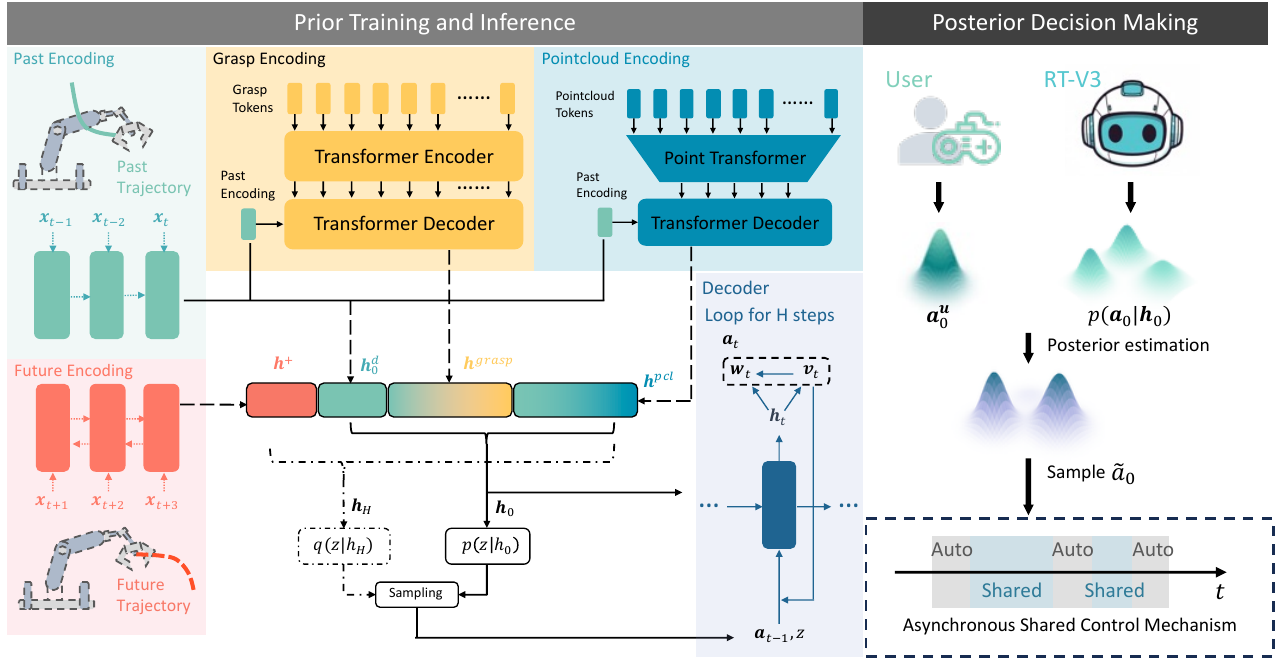}
  \caption{The overall architecture of Robot Trajectron V3 (RT-V3). \textbf{[Left] Prior Training and Inference:} The model extracts representations from past trajectories, candidate grasp poses, and point clouds to form the context embedding $\bm{h}_0$. Following a CVAE framework, the maneuver class $z$ is sampled and fed into an LSTM decoder, where the action generation is explicitly factorized into a translational velocity $v_t$ and a translation-conditioned rotational velocity $w_t$. \textbf{[Right] Posterior Decision Making:} During teleoperation, RT-V3 performs Bayesian posterior estimation, fusing the uncertain user command $\bm{a}^u_0$ with the learned behavioral prior $p(\bm{a}_0|\bm{h}_0)$ at the current timestep. Furthermore, an asynchronous shared control mechanism dynamically alternates between shared autonomy and autonomous execution to reduce user cognitive fatigue.}
  \label{fig: rtv3}
\end{figure*}

\section{$SE(3)$ Grasp-conditioned Policy} \label{sec. arch}
In this section, we detail the parameterization of the prior model that clones the user's behavioral patterns, described in Eq.~\ref{eq: decomposed mle with latent}. The proposed RT-V3's framework is shown in Fig.~\ref{fig: rtv3}. We assume a static environment in which the context does not change. With this assumption, we can factorize the past state trajectory $\bm{s}_{-T+1:0}$ into a past dynamic trajectory $\bm{x}_{-T+1:0}$ and a static context $\bm{c}$, which consists of the point clouds of the scene $\mathcal{P}$ and the candidate destinations $\mathcal{G}$. Thus, $\bm{h}_0$ in Eq.~\ref{eq: decomposed mle with latent} can be factorized as $[\bm{h}^\textnormal{d}_0, \bm{h}^\textnormal{c}]$, and $\bm{h}_H$ can be factorized as $[\bm{h}^\textnormal{d}_0, \bm{h}^{+}, \bm{h}^\textnormal{c}]$ in Eq.~\ref{eq: ELBO}, where $\bm{h}^\textnormal{d}_0$ is the past dynamic encoding, $\bm{h}^\textnormal{c}$ is the contextual encoding, and $\bm{h}^{+}$ is the future encoding. The following subsections detail the extraction of these encodings.

All poses, velocities, and point cloud observations are expressed in the coordinate frame of the current end-effector pose. This egocentric representation reduces variability in the data and implicitly enforces $SE(3)$-equivariance of the learned policy. The proof for $SE(3)$-equivariance can be found in the Appendix.

\subsection{Dynamics Encoding}
Given the past dynamic trajectory $\bm{x}_{-T+1:0}$, where $\bm{x}_t = [X_{t}, \dot{X}_{t}] \in \mathbb{R}^{12}$ is the dynamics information of the robot (i.e., the pose and the twist of the end-effector expressed in the Lie algebra), we use LSTM to obtain the past dynamics encoding $\bm{h}^{\textnormal{d}}_0$, as:
\begin{equation}
    \bm{h}^{\textnormal{d}}_0 = \textnormal{LSTM}(\bm{x}_{-T+1:0}).
\end{equation}
Additionally, the future dynamics encoding $\bm{h}^{+}$ is obtained by using BiLSTM to encode $\bm{x}_{1:H}$, as:
\begin{equation}
    \bm{h}^{+} = \textnormal{BiLSTM}(\bm{x}_{1:H}).
\end{equation}

\subsection{Context Encoding}
RT-V3 encodes two sources of context: point clouds $\mathcal{P}$ and candidate grasps $\mathcal{G}$. The feasible grasps can be obtained from off-the-shelf grasp planners \cite{song2025equivariant}. We aim to obtain a contextual encoding $\bm{h}^{\textnormal{c}}$, which consists of the point cloud encoding $\bm{h}^{\textnormal{pcl}}$ and the grasp encoding $\bm{h}^{\textnormal{grasp}}$.

\noindent \textbf{Point cloud encoding}. When the user controls the robot, they tend to move the robot to the desired position without a collision with objects in the workspace. Thus, the point clouds provide information about how the user would move the robot. As shown in Fig.~\ref{fig: rtv3}, we use PT-V3 \cite{wu2024point} as an encoder to preliminarily encode the point clouds, as:
\begin{equation}
    \bm{f}^{\textnormal{pcl}}=\mathop{\textnormal{PT-V3}}(\mathcal{P}),
\end{equation}
where $\bm{f}^{\textnormal{pcl}} \in \mathbb{R}^{\tilde{M}\times C_{\textnormal{p}}}$, $\tilde{M}$ is the number of points in the pooled point clouds, and $C_\textnormal{p}$ is the number of the encoded feature's channels. The initial point feature $\bm{f}^{\textnormal{pcl}}$ summarizes the object geometry of the workspace. We propose to use a transformer decoder to decode $\bm{h}^{\textnormal{pcl}}$ with the past dynamics encoding as a query, and decode the point cloud encoding as:
\begin{equation}
    \bm{h}^{\textnormal{pcl}}=\mathop{\textnormal{TransformerDecoder}}(\bm{f}^{\textnormal{pcl}},\bm{h}_0^{\textnormal{d}}).
\end{equation}
Powered by the cross-attention mechanism in the transformer decoder, we can dynamically extract the geometry information according to the current dynamics.

\noindent \textbf{Grasp encoding}. We assume that the dense set of candidate grasp poses from the off-the-shelf grasp planner includes the grasp poses that the user desires to execute. Thus, providing the model with grasp information is essential for predicting the future trajectories. Similar to how point cloud encoding is obtained, we use a transformer encoder to encode the grasps, followed by a transformer decoder with the past dynamics encoding as a query, as:
\begin{align}
    & \bm{f}^{\textnormal{grasp}}=\mathop{\textnormal{TransformerEncoder}}(\mathcal{G})\\
    & \bm{h}^{\textnormal{grasp}}=\mathop{\textnormal{TransformerDecoder}}(\bm{f}^{\textnormal{grasp}},\bm{h}_0^{\textnormal{d}}).
\end{align}
Similarly, the cross-attention mechanism in the transformer decoder allows the model to adaptively focus on the most probable grasps and predict trajectories towards them.
Finally, we obtain the context encoding $\bm{h}^{\textnormal{c}}=[\bm{h}^{\textnormal{pcl}};\bm{h}^{\textnormal{grasp}}]$.

\subsection{Maneuver Class Sampling}
We model maneuver class prior $p(\bm{z}|\bm{h}_0)$ and posterior $q(\bm{z}|\bm{h}_H)$ as categorical distributions whose parameters are generated with multi-layer perceptrons (MLPs), as:
\begin{align}
    & B_{\textnormal{prior}} = \textnormal{MLP}(\bm{h}_{0}),\\
    & B_\textnormal{posterior} = \textnormal{MLP}(\bm{h}_{H}).
\end{align}
We sample $\bm{z}$ from $p(\bm{z}|\bm{h}_0)$ during the training, while from $q(\bm{z}|\bm{h}_H)$ during the inference.

\subsection{Decoding Future Trajectory}
When decoding the future trajectory, we use an LSTM model to encode the latent dynamics, as:
\begin{equation}
    \bm{h}_{t+1} = \textnormal{LSTM}_t(\bm{h}_{t},\bm{a}_{t},\bm{z}).
\end{equation}
The action $\bm{a}_t$ is an $\mathfrak{se}(3)$ twist $[\bm{v}_t, \bm{w}_t] \in \mathbb{R}^6$, a concatenation of translational and rotational velocity. Modeling $\mathfrak{se}(3)$ actions as a joint 6D distribution is challenging due to the high dimensionality and strong coupling between translational and rotational components. Moreover, in many manipulation tasks, translational motion determines the global trajectory structure, while orientation adjustments are typically performed as local refinements near the target pose. Motivated by these observations, we factorize the action distribution $p(\bm{a}_t|\bm{h}_t, \bm{z})$ into a translational component $p(\bm{v}_t|\bm{h}_t, \bm{z})$  and a rotation component conditioned on translation $p(\bm{w}_t|\bm{h}_t, \bm{z},\bm{v}_t)$, as:
\begin{equation}
    p(\bm{a}_t|\bm{h}_t, \bm{z})=p(\bm{w}_t|\bm{h}_t, \bm{z},\bm{v}_t)p(\bm{v}_t|\bm{h}_t, \bm{z}). \label{eq: factorization 1}
\end{equation}
This design simplifies density estimation and improves stability during closed-loop execution. 

Specifically, we model the translational distribution $p(\bm{v}_t|\bm{h}_t, \bm{z})$ as a Gaussian distribution. For a given maneuver class $\bm{z}$, the parameters $G^\textnormal{v}_t(\bm{z}) = (\bm{\mu}_t^{z}, \bm{\Sigma}_t^{z})$ are generated via an MLP:
\begin{equation}
    (\bm{\mu}_t^{z}, \bm{\Sigma}_t^{z}) = \textnormal{MLP}(\bm{h}_{t}, \bm{z}).  \label{eq:velocitygmm}
\end{equation}
The translational action at time $t$ is then sampled as $\bm{v}^{z}_{t} \sim \mathcal{N}(\bm{\mu}_t^{z}, \bm{\Sigma}_t^{z})$.

Similarly, the rotational distribution $p(\bm{w}_t|\bm{h}_t, \bm{z}, \bm{v}_t)$ is modeled as a Gaussian conditioned on the sampled translation $\bm{v}_t$. The parameters $G^\textnormal{w}_t(\bm{z}) = (\bm{\omega}_t^{z}, \bm{\Gamma}_t^{z})$ are decoded as:
\begin{equation}
    (\bm{\omega}_t^{z}, \bm{\Gamma}_t^{z}) = \textnormal{MLP}(\bm{h}_{t}, \bm{z}, \bm{v}_t).  \label{eq:rotgmm}
\end{equation}
The rotational action is subsequently sampled as $\bm{w}^{z}_{t} \sim \mathcal{N}(\bm{\omega}_t^{z}, \bm{\Gamma}_t^{z})$. By marginalizing over the categorical maneuver class $p(\bm{z}|\bm{h}_{0})$, the joint action space implicitly forms a Gaussian Mixture Model (GMM), capturing the highly multi-modal nature of user intent.

During inference, RT-V3 sequentially samples $\bm{z} \sim p(\bm{z}|\bm{h}_{0})$, followed by $\bm{v}_t \sim p(\bm{v}_t|\bm{h}_t, \bm{z})$ and $\bm{w}_t \sim p(\bm{w}_t|\bm{h}_t, \bm{z}, \bm{v}_t)$ at each timestep. The final $SE(3)$ pose trajectory is integrated using rigid body dynamics:
\begin{equation}
T_t = T_{t-1} \mathop{\textnormal{Exp}}(\bm{a}_t \Delta t),
\end{equation}
where $T_t$ represents the $SE(3)$ transformation matrix at timestep $t$, and $\mathop{\textnormal{Exp}}(\cdot)$ denotes the Lie group exponential map.

\section{Posterior Decision Making} \label{posterior}
Until now, we have focused exclusively on modeling the user's behavior with a prior model $p(\bm{h}_{0:H},\bm{a}_{0:H-1}|\bm{s}_{-T+1:0})$, and the parametrization of this model. In this section, we expand the prior model in Sec.~\ref{sec. psc} to consider user input, in addition to historical motion and contextual cues.


At the current timestep, we receive the user command $\bm{a}^{\text{u}}_{0}=[\bm{v}_0^{\textnormal{u}}, \bm{w}_0^{\textnormal{u}}]$ from an uncertain interface. To achieve a better informed estimation of the intended action sequence $\bm{a}_{0:H-1}$, we incorporate the observed user command $\bm{a}^{\text{u}}_{0}$ and infer the posterior decision-making distribution $p(\bm{h}_{0:H}, \bm{a}_{0:H-1}|\bm{s}_{-T+1:0}, \bm{a}^{\text{u}}_{0})$. Following the same technique in Eq.~\ref{eq: decomposed mle with latent}, we can decompose Eq.~\ref{eq: posterior decision making} as:
\begin{equation}
\begin{aligned}
     & p(\bm{h}_{0:H}, \bm{a}_{0:H-1}|\bm{s}_{-T+1:0}, \bm{a}^{\text{u}}_{0}) \\
      &  = p(\bm{h}_{0}|\bm{s}_{-T+1:0})\\
      &~~ \sum_{\bm{z}}~p(\bm{h}_{1}|\bm{h}_{0}, \bm{a}_{0}, \bm{z}) \mathcolorbox{blue!10}{p(\bm{a}_{0}|\bm{h}_{0},\bm{z}, \bm{a}_0^{\textnormal{u}})}
     \mathcolorbox{green!15}{p(\bm{z}|\bm{h}_{0}, \bm{a}_0^{\textnormal{u}})} \\
     & ~~\cdot \prod_{t=1}^{H-1} p(\bm{h}_{t+1}|\bm{h}_{t}, \bm{a}_{t}, \bm{z}) p(\bm{a}_{t}|\bm{h}_{t},\bm{z}).
    \label{eq: shared control goal}
\end{aligned}
\end{equation}
Compared to Eq.~\ref{eq: decomposed mle with latent}, the terms $\mathcolorbox{blue!10}{p(\bm{a}_0|\bm{h}_{0},\bm{z})}$ and $\mathcolorbox{green!15}{p(\bm{z}|\bm{h}_{0})}$ become $\mathcolorbox{blue!10}{p(\bm{a}_{0}|\bm{h}_{0},\bm{z}, \bm{a}_0^{\textnormal{u}})}$ and $\mathcolorbox{green!15}{p(\bm{z}|\bm{h}_{0}, \bm{a}_0^{\textnormal{u}})}$ in Eq.~\ref{eq: shared control goal}, respectively. The remaining terms in Eq.~\ref{eq: shared control goal} are identical to those in Eq.~\ref{eq: decomposed mle with latent}. To compute $p(\bm{h}_{0:H}, \bm{a}_{0:H-1}|\bm{s}_{-T+1:0}, \bm{a}_0^{\textnormal{u}})$, we must first calculate $\mathcolorbox{blue!10}{p(\bm{a}_{0}|\bm{h}_{0},\bm{z}, \bm{a}_0^{\textnormal{u}})}$ and $\mathcolorbox{green!15}{p(\bm{z}|\bm{h}_{0}, \bm{a}_0^{\textnormal{u}})}$. 

Following the formulation in Eq.~\ref{eq: factorization 1}, we can factorize $\mathcolorbox{blue!10}{p(\bm{a}_{0}|\bm{h}_{0},\bm{z}, \bm{a}_0^{\textnormal{u}})}$ into a translational distribution and a rotational distribution conditioned on the translation:
\begin{equation}
\begin{aligned}
   & \mathcolorbox{blue!10}{p(\bm{a}_{0}|\bm{h}_{0},\bm{z}, \bm{a}_0^{\textnormal{u}})}= p(\bm{v}_0, \bm{w}_{0}|\bm{h}_{0},\bm{z}, \bm{v}_0^{\textnormal{u}}, \bm{w}_0^{\textnormal{u}}),\\
   & = \mathcolorbox{lime!20}{p(\bm{v}_{0}|\bm{h}_{0}, \bm{z}, \bm{v}_0^{\textnormal{u}})} \mathcolorbox{lightblue!70}{p(\bm{w}_{0}|\bm{h}_{0}, \bm{z}, \bm{v}_0, \bm{w}_0^{\textnormal{u}})} 
\end{aligned}
\end{equation}
Based on Bayesian rule, we can express $\mathcolorbox{lime!20}{p(\bm{v}_{0}|\bm{h}_{0}, \bm{z}, \bm{v}_0^{\textnormal{u}})}$ as the product of the prior and likelihood:
\begin{equation}
   \mathcolorbox{lime!20}{p(\bm{v}_{0}|\bm{h}_{0}, \bm{z}, \bm{v}_0^{\textnormal{u}})} \propto p(\bm{v}_{0}|\bm{h}_{0}, \bm{z}) p(\bm{v}_0^{\textnormal{u}}|\bm{v}_{0}). \label{eq: translation factorization}
\end{equation}
We assume $p(\bm{v}_0^{\text{u}}|\bm{v}_{0}) = \mathcal{N}(\bm{v}_0^{\text{u}}|\bm{v}_0, \bm{\Sigma}^{\text{sys}})$, where $\bm{\Sigma}^{\text{sys}}$ represents the system covariance of the user interface for the translational velocity. Then, we can analytically calculate $\mathcolorbox{lime!20}{p(\bm{v}_0|\bm{h}_{0}, \bm{z}, \bm{v}_0^{\text{u}})}$ as:
\begin{align}
    & \mathcolorbox{lime!20}{p(\bm{v}_0|\bm{h}_{0}, \bm{z}, \bm{v}_0^{\text{u}})} = \mathcal{N}(\bm{v}_0|\tilde{\bm{\mu}}_0^z, \tilde{\bm{\Sigma}}_0^z),\\
    & \bm{K}_{\text{v}} = \bm{\Sigma}_0^z(\bm{\Sigma}_0^z+\bm{\Sigma}^{\text{sys}})^{-1},\\
    & \tilde{\bm{\mu}}_0^z = \bm{\mu}_0^z + \bm{K}_{\text{v}}(\bm{v}_0^{\textnormal{u}}-\bm{\mu}_0^z), \\
    & \tilde{\bm{\Sigma}}_0^z = (\bm{I}-\bm{K}_{\text{v}})\bm{\Sigma}_0^z.
\end{align}
Similar to Eq.~\ref{eq: translation factorization}, we can also express $\mathcolorbox{lightblue!70}{p(\bm{w}_{0}|\bm{h}_{0}, \bm{z}, \bm{v}_0, \bm{w}_0^{\textnormal{u}})}$ as the
product of the prior and likelihood, as follows:
\begin{equation}
   \mathcolorbox{lightblue!70}{p(\bm{w}_{0}|\bm{h}_{0}, \bm{z}, \bm{v}_0, \bm{w}_0^{\textnormal{u}})} \propto p(\bm{w}_{0}|\bm{h}_{0}, \bm{z}, \bm{v}_0) p(\bm{w}_0^{\textnormal{u}}|\bm{w}_{0})
\end{equation}
By assuming $p(\bm{w}_0^{\text{u}}|\bm{w}_{0}) = \mathcal{N}(\bm{w}_0^{\text{u}}|\bm{w}_0, \bm{\Gamma}^{\text{sys}})$, where $\bm{\Gamma}^{\text{sys}}$ represents the system covariance of the user interface for the rotational velocity, we can obtain $\mathcolorbox{lightblue!70}{p(\bm{w}_{0}|\bm{h}_{0}, \bm{z}, \bm{v}_0, \bm{w}_0^{\textnormal{u}})}$ as:
\begin{align}
    & \mathcolorbox{lightblue!70}{p(\bm{w}_{0}|\bm{h}_{0}, \bm{z}, \bm{v}_0, \bm{w}_0^{\textnormal{u}})} = \mathcal{N}(\bm{w}_0|\tilde{\bm{\omega}}_0^z, \tilde{\bm{\Gamma}}_0^z),\\
    & \bm{K}_{\text{w}} = \bm{\Gamma}_0^z(\bm{\Gamma}_0^z+\bm{\Gamma}^{\text{sys}})^{-1},\\
    & \tilde{\bm{\omega}}_0^z = \bm{\omega}_0^z + \bm{K}_{\text{w}}(\bm{w}_0^{\textnormal{u}}-\bm{\omega}_0^z), \\
    & \tilde{\bm{\Gamma}}_0^z = (\bm{I}-\bm{K}_{\text{w}})\bm{\Gamma}_0^z.
\end{align}
For $\mathcolorbox{green!15}{p(\bm{z}|\bm{h}_{0}, \bm{a}_0^{\textnormal{u}})}$, we can factorize it with Bayesian rule, as:
\begin{equation}
     \mathcolorbox{green!15}{p(\bm{z}|\bm{h}_{0}, \bm{a}_0^{\text{u}})} = \frac{p(\bm{a}_0^{\text{u}}| \bm{h}_{0}, \bm{z})}{p(\bm{a}_0^{\text{u}}|\bm{h}_{0})} p(\bm{z}|\bm{h}_{0}).
\end{equation}
Since each maneuver class $\bm{z}$ corresponds to a component in action-GMMs, we can evaluate $p(\bm{a}_0^{\text{u}}| \bm{h}_{0}, \bm{z})$ and $p(\bm{a}_0^{\text{u}}|\bm{h}_{0})$ as follows:
\begin{equation}
     p(\bm{a}_0^{\text{u}}|\bm{h}_{0}, \bm{z}) = \mathcal{N}(\bm{v}_0^{\text{u}}|\bm{\mu}^{z}_0, \bm{\Sigma}^{z}_0) \mathcal{N}(\bm{w}_0^{\text{u}}|\bm{\omega}^{z}_0, \bm{\Gamma}^{z}_0)
\end{equation}
\begin{equation}
     p(\bm{a}_0^{\text{u}}|\bm{h}_{0}) = \sum_{\bm{z}} p(\bm{z}|\bm{h}_{0}) p(\bm{a}_0^{\text{u}}|\bm{h}_{0}, \bm{z}).
\end{equation}

The graphical model of the posterior decision making is represented in Fig.~\ref{fig: graphical model}~(b).
With $\mathcolorbox{green!15}{p(\bm{z}|\bm{h}_{0}, \bm{a}_0^{\textnormal{u}})}$ and $\mathcolorbox{blue!10}{p(\bm{a}_{0}|\bm{h}_{0},\bm{z}, \bm{a}_0^{\textnormal{u}})}$, we can derive the posterior distribution $p(\bm{h}_{0:H}, \bm{a}_{0:H-1}|\bm{s}_{-T+1:0}, \bm{a}_0^{\textnormal{u}})$.  This distribution yields a better-informed estimate of the intended action sequence $\{\bm{a}_t\}_{t=0}^{H-1}$, incorporating the current user's command $\bm{a}^{\textnormal{u}}_0$.

\noindent \textbf{Asynchronous Shared Control Mechanism.}  While RT-V3 maintains high-frequency inference (e.g., 20 Hz), real-world teleoperation often involves intermittent user input as operators pause to reassess their strategy. In RT-V2 \cite{rtv2}, the robot remains idle if the user stops providing commands. To address this, we propose an asynchronous shared control framework that leverages RT-V3’s capacity for autonomous action execution during periods of user inactivity, thereby reducing cognitive load. Specifically, the system monitors user inputs at each timestep: if a command is detected, RT-V3 performs posterior decision-making; otherwise, it executes the most probable action from its autonomous policy. Consequently, as long as the robot’s motion aligns with the user's intent, the operator can refrain from continuous signaling.

To prevent autonomous execution from interfering with fine-grained user adjustments during grasp acquisition, this mechanism is disabled when the end-effector is within a predefined distance of the object surface. At close proximity, operators often perform subtle corrective motions to refine the final grasp pose, whereas continued autonomous motion based on previously inferred intent may hinder such corrections. Disabling asynchronous execution in this region ensures that robot motion remains tightly coupled to ongoing user inputs during the final approach phase, while retaining the efficiency benefits of autonomous assistance during the earlier coarse-motion stages of the task.

As demonstrated in our user study, this mechanism substantially reduces the frequency of user commands and correspondingly lowers user effort without compromising task performance.


\section{Experiment}
We evaluate the proposed RT-V3 framework through extensive experiments in both simulation and real-world settings. Our evaluation focuses on three core capabilities: intent prediction, reactive planning, and shared control. 
\begin{itemize} 
\item \textbf{Intent Prediction} (Sec.~\ref{sec: traj-pred}): We assess trajectory forecasting accuracy by measuring the discrepancy between predicted trajectories and the ground truth within our curated dataset. 
\item \textbf{Planning} (Sec.~\ref{sec: planning}): We evaluate the model's ability to reach a specified target grasp pose reactively in simulation environments. \item \textbf{Shared Control} (Sec.~\ref{sec: sim-shared} and \ref{sec: real-world exp}): To demonstrate generalizability, we first conduct large-scale testing with diverse simulated agents, followed by real-world trials with human users. These experiments validate RT-V3's assistive ability across varied user behaviors and environmental contexts. 
\end{itemize}

\begin{figure}[!t]
  \centering
  \includegraphics[width=\linewidth]{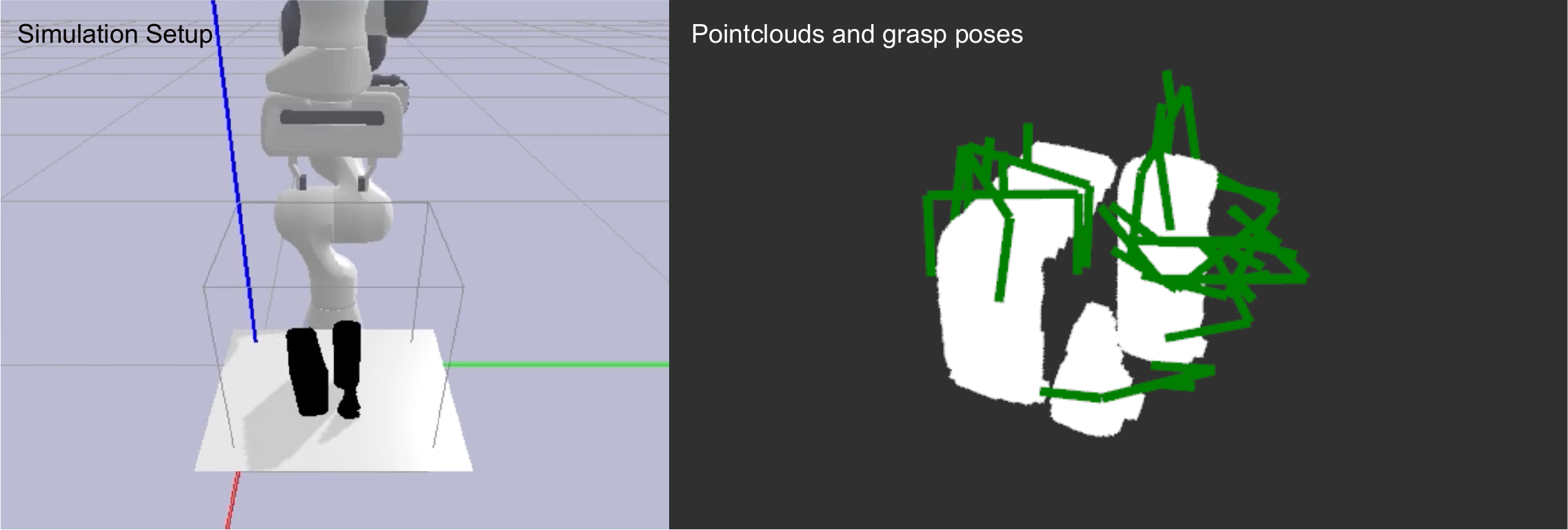}
  \caption{Simulation experiment setup. \textbf{[Left]} Simulation setup with a $30 \times 30 \times 30$ cm$^3$. \textbf{[Right]} The sampled point clouds and candidate grasp poses of the scene.}
  \label{fig: sim_setup}
\end{figure}

\subsection{Trajectory Prediction} \label{sec: traj-pred}
We first evaluate RT-V3's predictive performance using a simulation dataset. Following established benchmarks in trajectory forecasting \cite{mid,trajectron}, we employ two standard metrics: \begin{itemize} 
\item \textbf{Average Displacement Error (ADE)}: The mean manifold distance (translational Euclidean distance + scaled Euler angle error, see details in Appendix) between the ground truth and predicted positions over the entire trajectory. 
\item \textbf{Final Displacement Error (FDE)}: The manifold distance between the final predicted position and the ground truth endpoint. 
\end{itemize} 
To evaluate how effectively RT-V3's CVAE-GMM framework captures multi-modal user intent, we report both Best-of-Mode (BoM) and Most Likely (ML) metrics. BoM reflects the model's highest accuracy among all generated Gaussian modes, while ML measures the accuracy of the model’s most likely single trajectory.

\noindent \textbf{Simulation Setup and Data Collection.}
To train RT-V3, we curated a large-scale simulation dataset comprising point clouds, feasible grasp poses, and motion trajectories. The data generation pipeline consists of two stages:

\noindent \textbf{(i) Scene Generation}: Following the protocols of volumetric grasping models \cite{song2024implicit, song2025equivariant}, we utilized a set of 303 unique objects. Scenes were synthesized by randomly placing or dropping a subset of these objects into a $30 \times 30 \times 30$ cm$^3$ tabletop workspace (Fig.~\ref{fig: sim_setup}). For each scene, a depth map was captured from a randomized camera viewpoint to produce the corresponding point cloud. Feasible grasps were identified by sampling 6 rotation angles for each candidate approach vector and gripper width, resulting in a total of 833,324 valid grasps across 22,265 scenes.

\noindent \textbf{(ii) Trajectory Generation}: Using a Franka Research 3 robot as the experimental embodiment, we generated motion data. For each feasible grasp, a collision-free trajectory was computed from a random initial pose to the target using CuRobo \cite{curobo}. These trajectories were executed via a PD controller; we discarded any instances involving collisions, out-of-joint-limit, or failed grasps. The final dataset consists of 568,277 high-quality trajectories.

\noindent \textbf{Performance in the simulation dataset.} 
The results are presented in Tab.~\ref{tab:traj}. We conducted an ablation study to analyze the contribution of different components in RT-V3. The baseline model (first row) yields the highest errors in both BoM and ML metrics. 
The introduction of the \textbf{grasp encoder} provides the most substantial performance leap, further improving both BoM and ML performance (ADE-BoM 26.95, FDE-BoM 38.12, ADE-ML 30.36, and FDE-ML 44.17). This suggests that explicit goal-conditioning (grasp information) is crucial for accurate intent prediction. Finally, incorporating the \textbf{point cloud encoder} refines the prediction further, achieving a better BoM performance (ADE 26.70, FDE 37.90), which indicates that environmental geometric awareness helps in generating more feasible trajectories. Though adding point cloud information slightly reduces the ML performance, it helps improve planning performance, which will be explained in the next experiment. Additionally, we implement an RT-V3 variant that directly generates translation and rotation actions simultaneously (dubbed ``6D pred''), rather than using a translation-conditioned rotation architecture. It achieves the best performance in both BoM and ML settings (ADE-BoM 21.05, FDE-BoM 34.55, ADE-ML 22.50, and FDE-ML 37.65). However, we do not adopt this architecture due to its inferior planning performance; this limitation is analyzed further in the next experiment.

\begin{table*}[!t]
  \centering
  \caption{Performance of trajectory prediction and planning. Trajectory prediction metrics include average displacement error (ADE), final displacement error (FDE) in both Best-of-
  Mode (BoM) and Most Likely (ML) settings. Planning metrics include success rate (SR), collision rate (CR), and time-out rate (TOR). \bestt{Green} indicates best, and \secondt{blue} indicates second best.}
  
  \resizebox{0.9\textwidth}{!}{%
  
      \renewcommand{\arraystretch}{1.2} 
      \setlength{\tabcolsep}{3pt} 
      
      \begin{tabular}{cccccccccc} 
        \toprule
        \multicolumn{3}{c}{Method Components} & 
        \multicolumn{4}{c}{Trajectory Prediction Metrics ($\downarrow$)} & 
        \multicolumn{3}{c}{Planning Metrics} \\
        
        \cmidrule(lr){1-3} \cmidrule(lr){4-7} \cmidrule(lr){8-10}
        
        {Trans-Con Rot}  & {Grasp Enc.} & {PCL Enc.} & 
        {ADE-BoM} & {FDE-BoM} &  {ADE-ML} & {FDE-ML} & 
        SR ($\uparrow$) & CR ($\downarrow$) & TOR ($\downarrow$) \\
        \midrule
        
        \cmark &  & & 
        42.71 & 83.79 & 46.40 & 92.89 & 
        - & - & - \\
        
        
        \cmark  & \cmark & & 
        26.95 & 38.12 & \second{30.36} & \second{44.17} & 
        78.8 & 10.3 & \second{10.9} \\
        
        \cmark  & \cmark & \cmark & 
        \second{26.70} & \second{37.90} & 32.05 & 47.87 & 
        \second{90.4} & 7.9 & \best{1.7} \\
        
        & \cmark  & \cmark & 
        \best{21.05} & \best{34.55} & \best{22.50} & \best{37.65} & 
        3.4 & 68.7 & 27.9 \\
        
        \midrule
        
        \multicolumn{3}{c}{CuRobo \cite{curobo}} & 
        - & - & - & - & 
        \best{95.9} & \best{0} & - \\
        
        \multicolumn{3}{c}{NeoSS \cite{zhu2025efficient}} & 
        - & - & - & - & 
        83.5 & \second{5.4} & 11.1 \\
        
        \bottomrule
      \end{tabular}%
  } 
  \label{tab:traj}
\end{table*}

\begin{figure*}[!t]
  \centering
  \includegraphics[width=\linewidth]{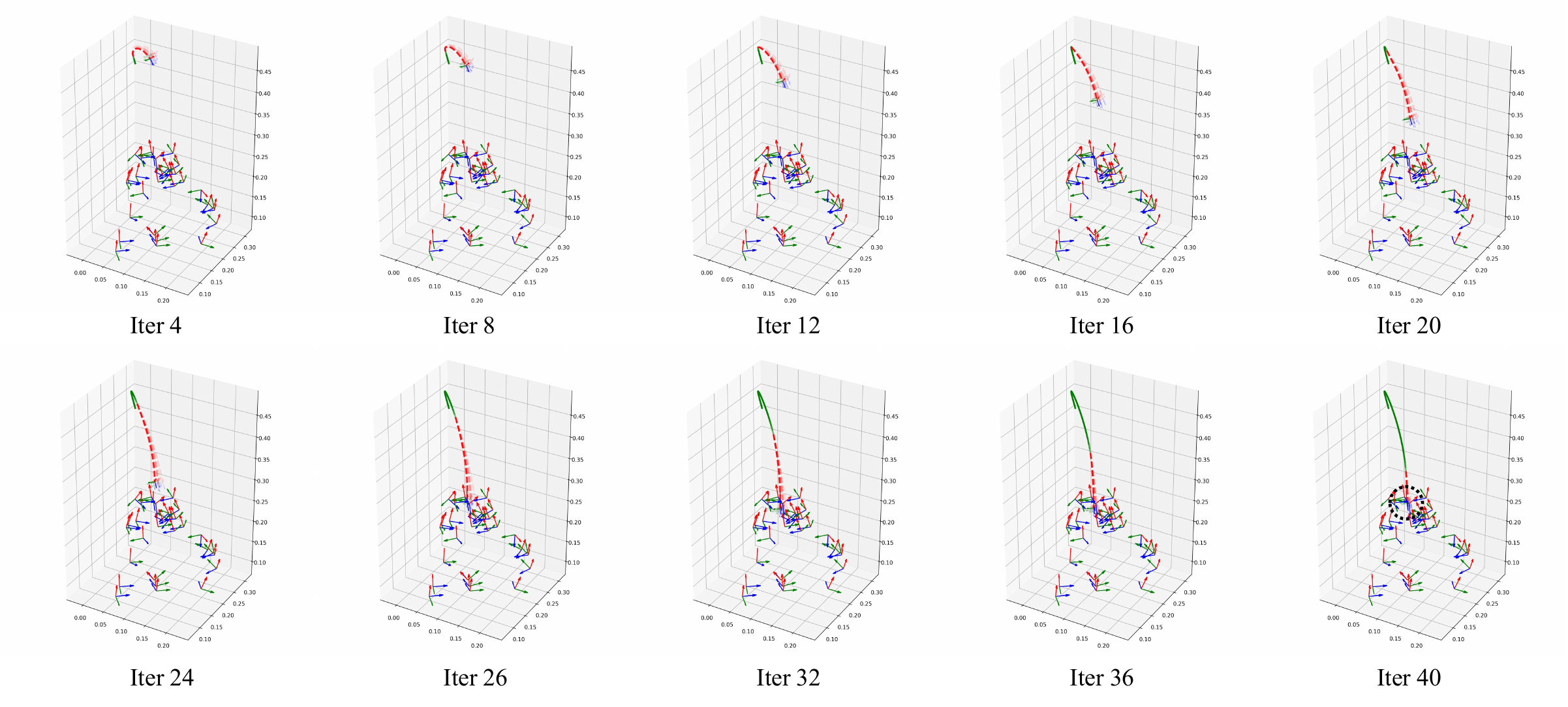}
  \caption{Visualization of RT-V3's predicted trajectories. For clarity, we visualize every four iterations. The past trajectories are painted as \textbf{\textcolor{377E22}{solid green lines}}, while the predicted trajectories are painted as \textbf{\textcolor{red}{dashed red lines}}. We also visualize the final step of the predicted trajectories and the grasp poses as frames. The opacity of the predicted trajectories denotes the value of $q_{\omega}({\bm{z}|\bm{h}_{0}})$. The black circle in iteration 40 highlights that the predicted trajectories reach the desired grasp poses.
   }
  \label{fig: trajpred}
\end{figure*}

\subsection{Planning in the Simulation Environment} \label{sec: planning}
RT-V3 itself is a reactive planning policy, so we aim to evaluate the planning ability of RT-V3 in this experiment. The planning experiment is conducted in a simulated environment identical to the one used for data collection in Sec.~\ref{sec: traj-pred}. We select two grasp poses in each scene as target poses. If RT-V3 can control the robot to reach one of the target poses without colliding with the table or objects, it will be regarded as a success, while failure occurs if the robot either collides with obstacles or fails to reach a goal within the time limit. The experiments are conducted over 1,000 trials with fixed random seeds, and three metrics are recorded: success rate (SR), collision rate (CR), and time-out rate (TOR). During each round, RT-V3 samples its optimal action using the ML setting as follows:
$ \bar{\bm{z}} = \mathop{\textnormal{argmax}}_{\bm{z}} p(\bm{z}|\bm{h}_{0}),
\bar{\bm{a}}_t = \mathop{\textnormal{argmax}}_{\bm{a}_t} p(\bm{a}_t|\bm{h}_t,\bar{\bm{z}})$.

\noindent \textbf{Planning performance.} We evaluated the model's reactive planning capability. RT-V3 (full version) achieves a 90.4\% Success Rate (SR), significantly outperforming the configuration without the PCL encoder (78.8\%). Notably, RT-V3 exhibits a lower time-out rate (TOR) (1.7\%) compared to the reactive controller NeoSS \cite{zhu2025efficient} (11.1\%), demonstrating its efficiency in reaching targets. Although RT-V3 underperforms CuRobo (95.9\% SR), a dedicated global planner, CuRobo does not support multi-modal intent prediction. Therefore, as an assistive controller, RT-V3 demonstrates competitive performance in planning. 

In contrast, the ``6D pred'' variant—which predicts translation and rotation simultaneously—exhibits poor planning performance, achieving only a 3.4\% SR. Because this variant is optimized primarily for open-loop trajectory prediction, it suffers from severe compounding errors (i.e., the "drift problem") during autonomous closed-loop execution, quickly driving the system into out-of-distribution states. By utilizing a translation-conditioned rotation architecture, the model avoids learning a complex 6D joint action distribution, opting instead for more tractable 3D conditional distributions. Furthermore, the sequential two-stage sampling process (sampling translation, followed by rotation conditioned on translation) introduces beneficial stochasticity during training. This noise acts as an effective regularizer, making the learned policy significantly more robust to distributional shifts during execution.

\subsection{Simulated Shared Control Experiments} \label{sec: sim-shared}
To evaluate the shared control capability of RT-V3, we propose a simulated shared control experiment. Inspired by the work of Reddy et al. \cite{reddy2018shared}, we propose to use simulated users and conduct shared control in the simulation environment, which can comprehensively evaluate RT-V3 at scale without using real human users. These simulated users are target-pose driven policies, but they reach the target with different behaviors. Specifically, we propose four different simulated users:
\begin{itemize}
    \item \textbf{Noisy User} generates the action twist towards the target pose with injected noises.
    \item \textbf{Laggy User} generates the action twist towards the target pose, but it is forced to repeat its previous action with probability 0.8.
    \item \textbf{Mode-Switching User} has two action modes: translation mode and rotation mode. In translation (or rotation) mode, the user can only generate translational (or rotational) action towards the target at a time. These two modes alternate every 10 seconds. 
    \item \textbf{Single-DoF User} can only control one DoF at a time. The user will switch to another DoF that has the largest distance every 5 seconds.
\end{itemize}

We pre-generate 50 validation scenes following the paradigm mentioned in Sec.~\ref{sec: traj-pred}, and sample 10 to 20 feasible grasps for each scene. In each trial, we set each grasp pose as the target pose for simulated users (a total of 768 trials).

\noindent \textbf{Experimental results.} Tab.~\ref{tab: sim shared} summarizes the performance of RT-V3 and several baseline methods across four types of simulated users. We record three metrics: success rate (SR), collision rate (CR), and the number of iterations used in the success trial (Iter).

We compare RT-V3 with two other baselines: pure teleoperation (``Direct'') and Hindsight Optimization (``HO''). RT-V3 outperforms them across all user types with higher SRs, lower CRs, and fewer iterations. HO does not improve the performance because the dense candidate poses tend to attract the robot along the way to its desired pose, trapping the robot in one of the candidate poses, leading to more iterations. Besides, HO does not have a planning capability, which makes it easy to collide with objects.

We also apply a collision avoidance method NeoSS \cite{zhu2025efficient} to Direct and HO. We observe that NeoSS largely decreases CR for both methods. Direct with NeoSS largely improves the SR in \emph{Noisy}, \emph{Mode-Switching}, and \emph{Single-Dof}, but their SRs are still lower than RT-V3. HO with NeoSS does not significantly improve the SR. These results demonstrate that to provide suitable assistance, the assistive algorithm must not only employ collision avoidance but also accurately estimate intent. The proposed RT-V3 combines planning and intent estimation in a unified Bayesian framework, achieving the highest SR across all user types. Combined with NeoSS, RT-V3 can achieve very low CR ($0\%\sim6\%$) with a small sacrifice of SR.

Besides, we conduct the ablated variant of RT-V3 ``6D pred'' in this shared control experiment, and its performances are inferior to the RT-V3, which further proves the effectiveness of the proposed translation-conditioned rotation architecture.

\begin{table*}[!t]
    \centering
    \caption{Shared control experiments across different user types. \bestt{Green} indicates best, and \secondt{blue} indicates second best.}
    \small 
    \setlength{\tabcolsep}{3pt} 
    \begin{tabular}{lcccccccccccc}
        \toprule
        \multirow{2}{*}{Method} & \multicolumn{3}{c}{Noisy} & \multicolumn{3}{c}{Laggy}  & \multicolumn{3}{c}{Mode-Switching} & \multicolumn{3}{c}{Single-Dof} \\
        \cmidrule(lr){2-4} \cmidrule(lr){5-7} \cmidrule(lr){8-10} \cmidrule(lr){11-13}
         & SR ($\uparrow$) & CR ($\downarrow$) & Iter ($\downarrow$) & SR ($\uparrow$) & CR ($\downarrow$) & Iter ($\downarrow$) & SR ($\uparrow$) & CR ($\downarrow$) & Iter ($\downarrow$) & SR ($\uparrow$) & CR ($\downarrow$) & Iter ($\downarrow$) \\
        \midrule
        Direct &  62.37 & 29.17 & 81.08 & 80.34 & 17.19 & 57.61  & 74.22 & 14.71 & 118.06 & 50.52 & 21.22 & 128.81 \\
        Direct w. NeoSS & 72.01 & \second{6.12} & 87.78 & 79.69 & \second{0.78} & 67.44 & \second{81.51} & \best{0.52} & 111.41 & 61.98 & \best{3.52} & 121.33 \\
        HO  & 42.84 & 31.64 & 103.02 & 9.37 & 13.41 & 143.65  & 17.97 & 15.23 & 136.56 & 23.05 & 21.61 & 131.24 \\
        HO w. NeoSS  & 49.09 & 11.59 & 107.22 & 9.90 & 3.52 & 142.72  & 17.57 & 4.04 & 137.45 & 25.52 & 7.03 & 131.51 \\
        \midrule
        RT-V3  & \best{84.11} & 11.07 & \second{68.37} & \best{86.06} & 9.77 & \second{50.93} & \best{83.33} & 4.68 & \second{105.40} & \best{69.79} & 10.81 & \second{107.17} \\
        RT-V3 w. NeoSS & \second{82.16} & \best{2.34} & 79.78 & \second{83.98} & \best{0.26} & 63.05 & 81.12 & \second{1.95} & 107.53 & \second{67.58} & \second{5.60} & 110.41 \\
        RT-V3 wo. Trans-Con Rot & 71.22 & 21.88 & \best{54.38} & 75.91 & 20.96 & \best{42.24} & 78.64 & 17.50 & \best{60.98} & 62.50 & 32.68 & \best{77.63} \\
        \bottomrule
    \end{tabular}
    \label{tab: sim shared}
\end{table*}

\begin{figure}[!t]
  \centering
  \includegraphics[width=\linewidth]{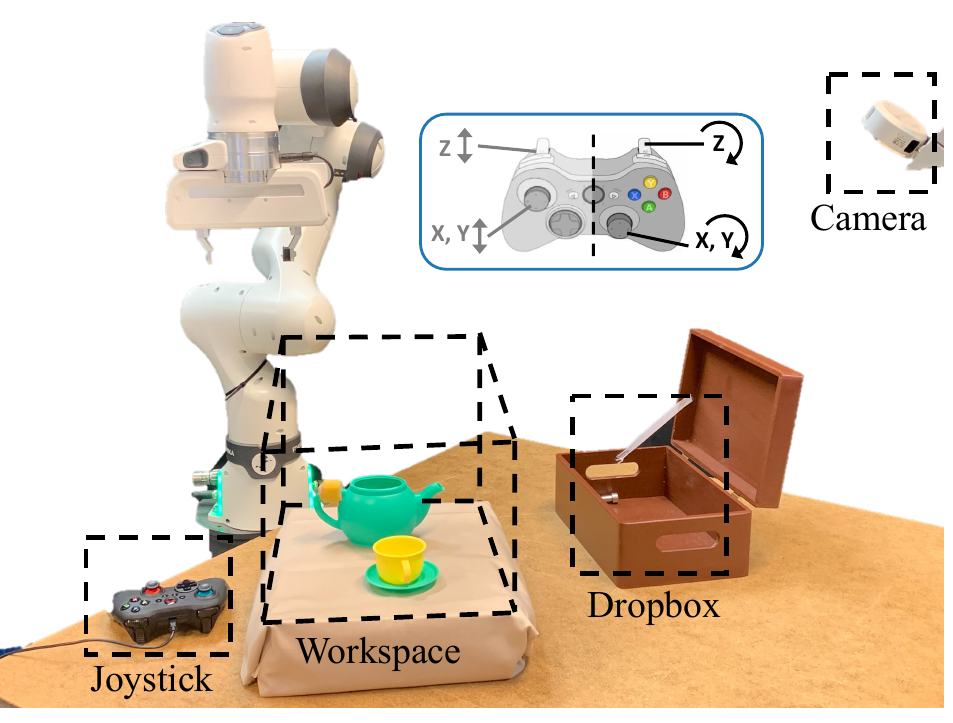}
  \caption{Real-world experimental setup with a Franka Research 3 robot and a top-down RGB-D. Users are required to control the robot to grasp the object in a certain way with an Xbox joystick.}
  \label{fig: setup}
\end{figure}

\subsection{Real-world Shared Control Experiments} \label{sec: real-world exp}
In this experiment, we aim to validate the performance of RT-V3 in a real-world shared autonomy setting. By having human users complete tasks in a real-world environment, assisted by RT-V3, we can assess its performance in terms of speed and user experience.

\noindent \textbf{Design.} In this experiment, the user will be asked to control the 7 DoF Franka Research robot with an XBox joystick to execute some grasping tasks. The left joystick, left trigger, and bumper are used to control the translation, and the right joystick, trigger, and bumper are used to control the rotation. ``A, B'' buttons are used to execute the grasp and release the gripper, respectively. The ``Y'' button is used to move along the Z-axis of the tool frame (dubbed ``approach grasp''), which is designed to move from pre-grasp to grasp poses. The ``X'' button is used to lift the object and drop it into the box once the user grasps the object. We compared the RT-V3 method against two baselines. The first baseline was pure user control (referred to as ``Direct''). The second baseline, Hindsight Optimization (HO) \cite{javdani2018shared}, is a well-known intent estimator. The experiment consisted of three rounds corresponding to a specific assistive method (or pure teleoperation), each comprising 4 trials. The order of the assistive methods is randomly shuffled for each user. Prior to each round, participants were instructed on the mechanics of the assistive algorithm and completed practice trials to mitigate potential learning effects. For each trial, two or three objects were placed within the workspace. Participants received a text command (e.g., ``grasp the facial cleanser from the right side'') and were tasked with retrieving the corresponding object. To mitigate learning effects, participants completed 12 unique trials across 12 distinct scenes. These scenes were divided into three subsets, with each subset randomly assigned to one of the three evaluated methods. A trial was classified as a failure under any of the following three conditions: (i) the robot experienced an unrecoverable collision or joint-limit violation; (ii) the execution exceeded the 90-second limit (``time-out''); or (iii) the robot disrupted the environment to an extent that rendered the task unachievable ('unmatched').

Before each round, we will explain to the user how this assistive algorithm works, and the user will practice with the algorithm on some tasks to get familiar with it in order to alleviate problems caused by proficiency. In each trial, 2 or 3 objects are placed in the workspace. The user will receive a text instruction, and they need to grasp the object in the workspace that matches the text instruction. For example, we may give the user an instruction like ``grasp the facial cleanser from the right side''. We prepared 12 different scenes for 12 trials, which can alleviate problems caused by proficiency. These 12 scenes were separated into 3 groups, and three methods were randomly assigned to one of the groups. A trial is considered a failure in three circumstances: (i) the robot encounters a collision or reaches the joint limit that makes it unable to recover; (ii) the operation time is out of 90 seconds (dubbed, ``out of time''); (iii) the robot destroys the scene and makes it impossible to match the text instruction (dubbed, ``unmatched'').

\noindent \textbf{Protocol.} We enrolled 21 novice participants from the local community (19 male, 2 female; age range: 18–40). The participants are able to control the robot's translation, rotation, grasping, and releasing. The order of the controllers was randomized for each participant to mitigate the effects of novelty and practice. After each round, participants rated their agreement with the following statements on a seven-point Likert scale:
\begin{itemize}
    \item This algorithm helped me complete the task \emph{quickly}.
    \item This robot did what I \emph{wanted}.
    \item This algorithm would be \emph{useful} for me in teleoperation.
    \item I felt in \emph{control} while using this algorithm.
    \item I would find this algorithm \emph{easy to use}.
    \item I \emph{trust} this algorithm.
\end{itemize}

Participants also completed a NASA-TLX survey \cite{nasatlx}, where they rated their workload across six subscales. \emph{Mental demand} assessed the cognitive and perceptual effort required; \emph{physical demand} measured the physical effort involved in the task; \emph{temporal demand} evaluated the time pressure experienced by the participant; \emph{effort} measured how hard the participant had to work to maintain performance; \emph{frustration} assessed feelings of annoyance, stress, and irritation; and \emph{performance} gauged how successful participants felt in completing the task. At the end of the three rounds, participants were also asked to indicate their preferred method and provide written comments.

All participants provided signed consent for the experimental procedure, which was approved by the Social and Societal Ethics Committee of (anonymous institution) (G-2024-8370). The collected data were processed in accordance with the General Data Protection Regulation (GDPR) of the (anonymous organization).

\noindent \textbf{Metrics.} Both objective and subjective metrics were used in the experiments. For objective metrics, we compared the success rate (SR), unmatched rate (UR), time-out rate (TOR), total joystick inputs, completion time, and both translational and rotational trajectory length. For subjective metrics, we compared participants' ratings from the agreement survey and their responses to the NASA-TLX survey across the three methods.

\begin{table}[t]
  \centering
  \caption{The success rate (SR), unmatched rate (UR), and time-out rate (TOR) of the real-world shared autonomy experiment. \bestt{Green} indicates best, and \secondt{blue} indicates second best.}
  \small 
  \begin{tabular}{cccc}
    \toprule
    Method & SR (\%) & UR (\%) & TOR (\%)\\
    \midrule
    Direct & \second{78.57}  & \second{16.67}  & \second{4.76} \\
    HO     & 74.70   & 18.07 & 7.23 \\
    RT-V3  &  \best{86.90}  & \best{9.53}  & \best{3.57}  \\
    \bottomrule
  \end{tabular}
  \vspace{-0.4cm}
  \label{tab:obj1}
\end{table}

\begin{figure*}[!t]
  \centering
  \includegraphics[width=\linewidth]{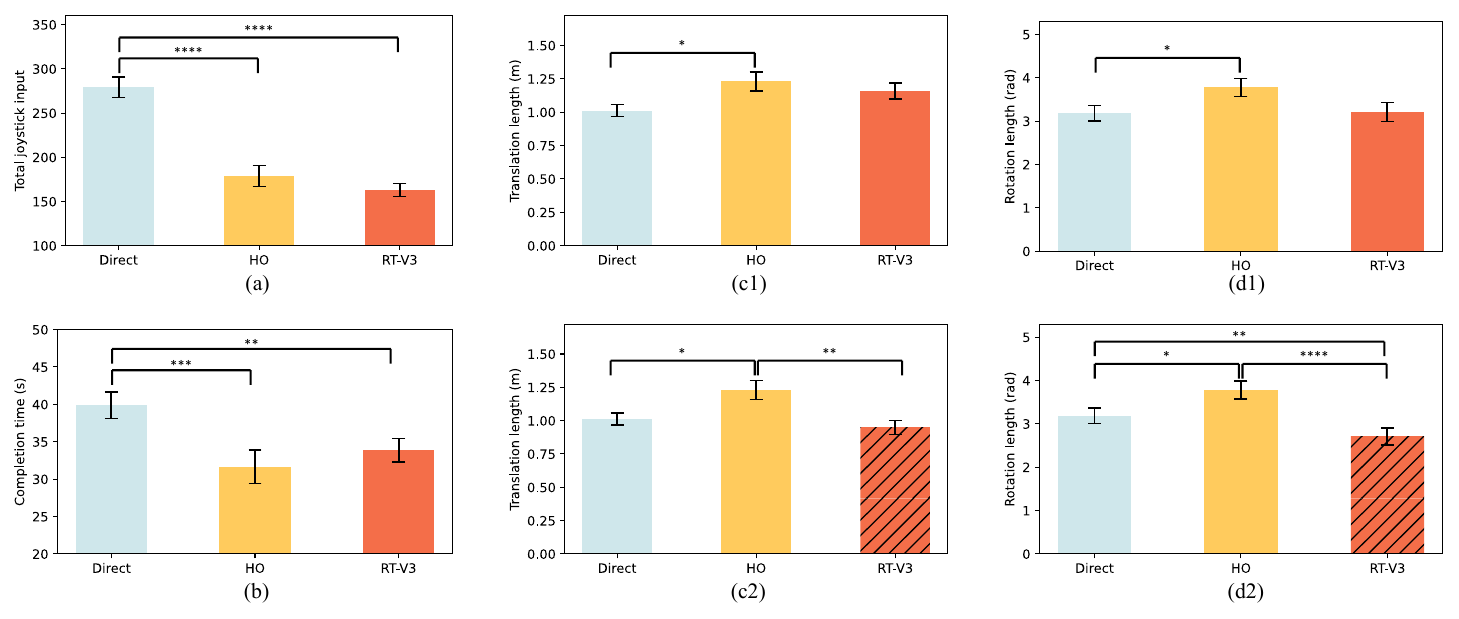}
  \caption{Some objective results for the real-world experiment. (a) Total joystick inputs. (b) Completion time. (c1) Translational trajectory length. (c2) Adjusted translational trajectory length. (d1) Adjusted rotational trajectory length. (d2) Rotational trajectory length.  $*= p < 0.05$, $**= p < 0.01$, and $***= p < 0.001$.}
  \label{fig: performance}
\end{figure*}

\noindent \textbf{Results of Objective Measures.} Tab.~\ref{tab:obj1} summarizes SR, UR, and TOR for each method. During the experiment, we did not encounter the first failure case (i.e., unrecoverable collision or out of joint limits). RT-V3 achieved the highest SR (86.90\%), the lowest UR (9.53\%), and the lowest TOR (3.57\%). HO easily trapped the robot in an undesired grasp pose, which is also observed in the simulated shared control experiment. This is because the performance of HO largely relies on the performance of the grasp planner. If the detected poses did not include the grasp pose the user desired, this ``trapped'' problem became more severe, leading to a high TOR of HO. Although RT-V3 also relies on the predicted grasps from the grasp planner, it is less constrained because it predicts the next action via posterior estimation instead of guiding the user towards the predicted intent pose. Therefore, RT-V3 also has the lowest TOR.

Fig.~\ref{fig: performance} compares \emph{total joystick input}, \emph{completion time}, and \emph{trajectory length}, excluding failure trials. The Kruskal-Wallis test was used to evaluate the effect of the method (Direct, HO, RT-V3) on these metrics. The Mann-Whitney test with a two-stage Benjamini-Hochberg post-hoc correction adjustment was applied for pairwise comparisons when significant effects were found.
For \emph{total joystick input} (Fig.~\ref{fig: performance}(a)), a significant difference was observed among the methods ($p<0.0001$), with RT-V3 and HO differing significantly from Direct ($p<0.0001$ and $p<0.0001$). Although the RT-V3's total joystick inputs are less than HO's, we did not observe a significant difference between RT-V3 and HO. 
For \emph{completion time} (Fig.~\ref{fig: performance}(b)), significant differences were found (\(p=0.0006\)), with both RT-V3 and HO differing from Direct ($p=0.01$ and $p=0.0012$). The HO's completion time is less than RT-V3's non-significantly due to the more aggressive nature of HO. 
For \emph{translation trajectory length} (Fig.~\ref{fig: performance}(c1), significant differences were found (\(p=0.0352\)), with Direct differing from HO significantly (\(p=0.0155\)). The translation length of Direct is shorter than RT-V3 non-significantly. We also observe similar effects in \emph{rotational trajectory length} where significant differences were found (\(p=0.0286\)). Direct's rotational length is significantly shorter than HO's (\(p=0.0155\)) and non-significantly shorter than RT-V3's, although Direct requires more joystick inputs and completion time. The user tends to control the robot degree by degree and sometimes stops to think about how to move the next step, but the length will not be largely increased. In contrast, assistive methods provide more aggressive control by moving all degrees of freedom simultaneously, but the user needs to spend extra trajectories on adjusting the robot when the intent prediction is incorrect, leading to a higher length of trajectory. Due to the asynchronous shared control mechanism, RT-V3 can automatically execute the predicted actions during the trial. Thus, we delete the trajectories when RT-V3 takes full control and plot the adjusted translation and rotation trajectory length in Fig.~\ref{fig: performance}(c2) and (d2). Significant differences were found in both adjusted translation and rotation trajectory length ($p=0.0009$ and $p<0.0001$). In the adjusted figure, the Direct's and RT-V3's translation and rotation lengths are significantly shorter than HO's. Notably, RT-V3's adjusted rotation length is significantly shorter than Direct's, which demonstrates RT-V3's superiority in assisting users in rotation manipulation.

\begin{figure*}[!t]
  \centering
  \includegraphics[width=\linewidth]{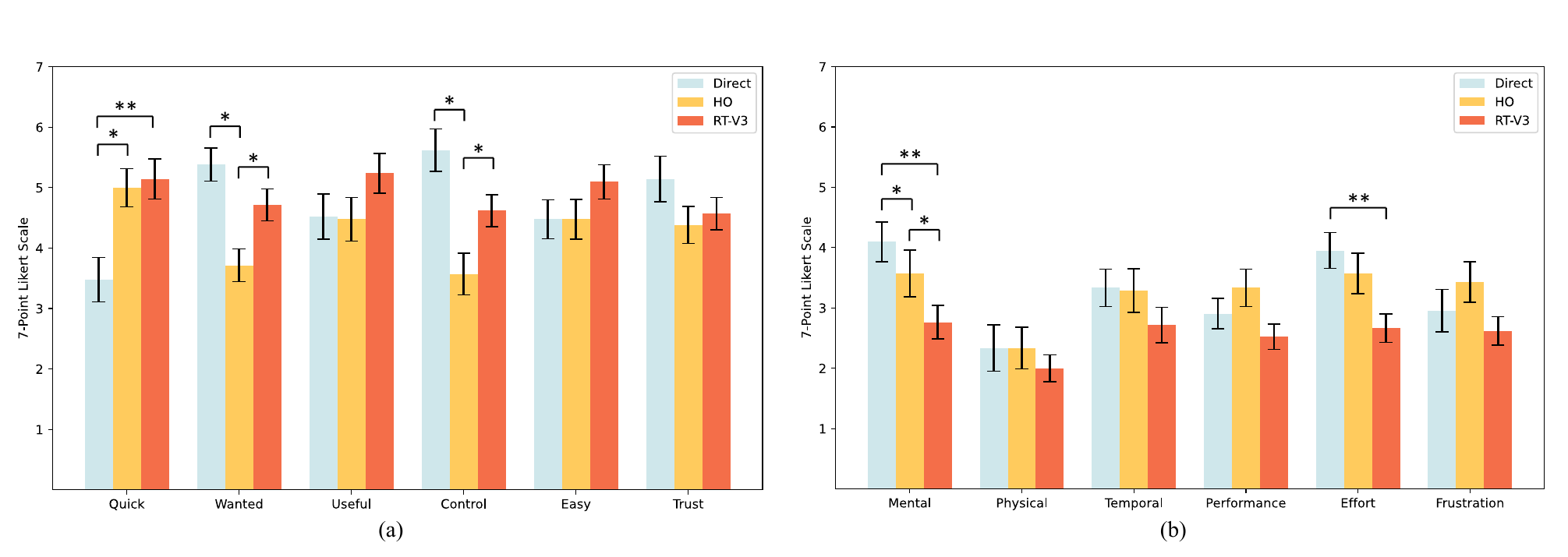}
  \caption{The user study for each method across all participants for the real-world shared control experiment. The plots are bar plots with error bars: (a) the agreement survey and (b) the NASA-TLX survey, where  $*= p < 0.05$ and $**= p < 0.01$. For (a) a higher score is better, while for (b) a lower score is better. The color of the bars representing the control methods is consistent across all figures.}
  \label{fig: likert}
\end{figure*}

\noindent \textbf{Results of Subjective Measures.} Fig.~\ref{fig: likert}~(a) illustrates the comparison of the agreement survey results. The Kruskal-Wallis H-test was employed to evaluate the effect of the method (Direct, HO, RT-V3) on these metrics. A Wilcoxon signed-rank test with a two-stage Benjamini-Hochberg correction was used for post hoc analysis of pairwise comparisons when significant effects were identified. Significant differences were observed in \emph{Quick}, \emph{Wanted}, and \emph{Control}. Post hoc analysis revealed significant differences between Direct and HO in \emph{Quick} ($p=0.0079$), \emph{Wanted} ($p=0.0139$), and \emph{Control} ($p=0.0113$). Additionally, significant differences were found between RT-V3 and Direct in \emph{Quick} ($p=0.0078$). Furthermore, RT-V3 differed significantly from HO in \emph{Wanted} ($p=0.0375$), \emph{Control} ($p=0.0113$). To summarize, RT-V3 is easier to use compared with HO because it gives users a sense of control, and RT-V3 makes users feel quicker compared with pure teleoperation by providing accurate assistance.

Fig.~\ref{fig: likert}~(b) demonstrates the comparison of the NASA-TLX survey. The Kruskal-Wallis H-test evaluated the effect of the method (Direct, HO, RT-V3) on these metrics. A Wilcoxon signed-rank test with two-stage Benjamini-Hochberg correction was used as post hoc analysis for pairwise comparisons when significant effects were found. RT-V3 obtains the best performance in all metrics. Significant differences were found for \emph{Mental} ($p=0.0268$), \emph{Effort} ($p=0.0174$). Post hoc analysis found significant differences between Direct and HO in \emph{Mental} ($p=0.0281$), between RT-V3 and Direct in \emph{Mental} ($p=0.0298$), \emph{Effort} ($p=0.0035$), and between RT-V3 and HO in \emph{Mental} ($p=0.0003$). No significant difference was found in \emph{Physical} demand, as the task involved only joystick control and was not physically taxing. Similarly, the lack of urgency in achieving the grasp poses resulted in no significant difference in \emph{Temporal} demand.

\begin{figure*}[!t]
  \centering
  \subfloat[Success case: ``Grasp the cap of the bottle from the left''.]{
    \includegraphics[width=\linewidth]{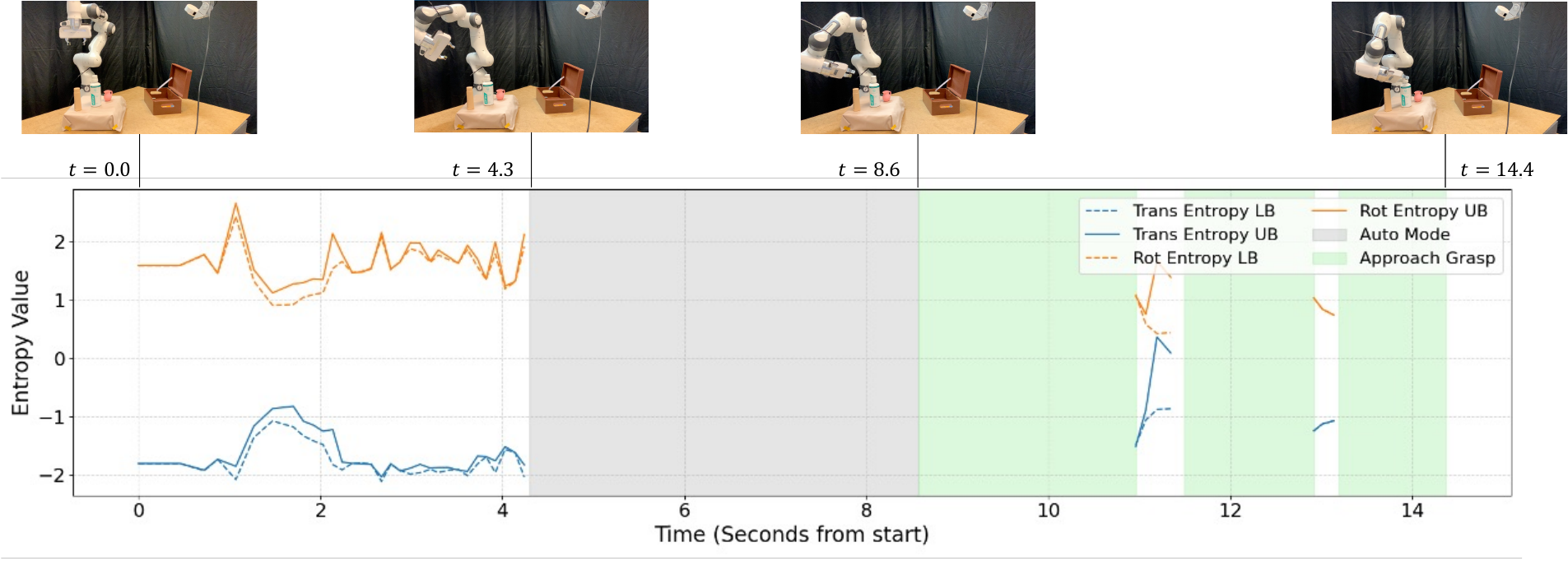}
    \label{fig:scene3_1_demo_sub} 
  }
  
  \vspace{0.5em} 
  
  \subfloat[Failure case: ``Grasp the pill bottle from the right''.]{
    \includegraphics[width=\linewidth]{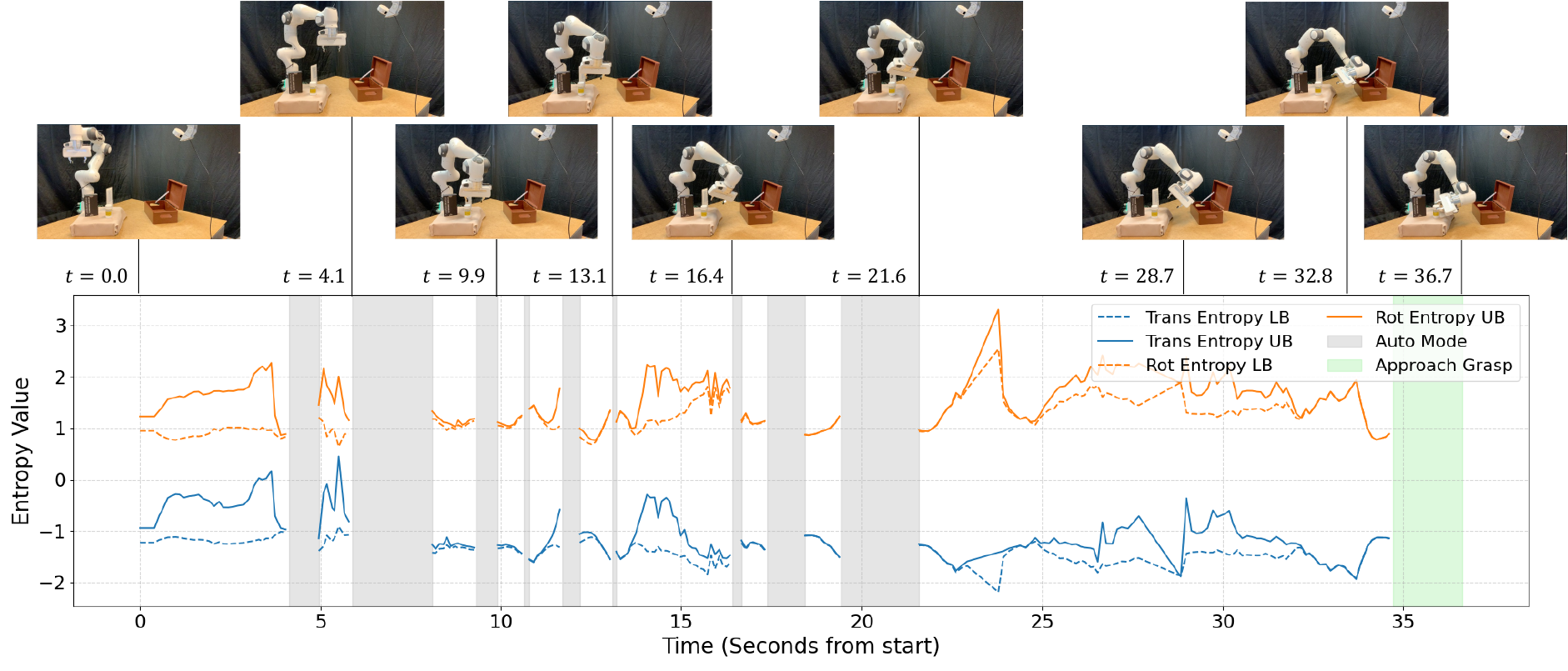}
    \label{fig:failure_sub} 
  }
  
  \caption{Visualization of task processes and entropy in successful and failed scenarios. 
  (a) A successful demonstration of grasping a bottle cap. 
  (b) A failure assistance case during the grasping of a pill bottle. 
  In both plots, the legend is consistent: \textcolor{blue!70}{Blue lines} denote the entropy of translational action distribution, and \textcolor{orange}{orange lines} denote the entropy of rotational action distribution. Solid lines denote the upper bound of the entropy, and dashed lines denote the lower bound. The \textcolor{gray}{gray zone} indicates autonomous control by RT-V3 using the prior, while the \textcolor{green!70}{green zone} denotes the \emph{approach grasp} action phase.}
  \label{fig:combined_entropy_vis} 
\end{figure*}

\noindent \textbf{Visualization of Guidance.} Fig.~\ref{fig:scene3_1_demo_sub} illustrates the real-time inference process and the corresponding uncertainty of the action distribution during a manipulation task. The text instruction is ``Grasp the cap of the bottle from the left''. At the beginning ($t=0.0$ to $4.3$), the robot was under shared control. The translational entropy (blue) and rotational entropy (orange) exhibited fluctuations as the model inferred user intent from multi-modal candidate grasp poses.
Specifically, the rotational entropy showed a distinct peak around $t=1.0$ and $2.8$, reflecting the ambiguity in selecting the optimal orientation among several candidate grasps.
From $t=4.3$ to $8.6$ (gray zone), the system entered an autonomous mode where RT-V3 predicted the future actions based on the past trajectories, and controlled the robot using the prior action distribution towards the grasp pose that it thought was most likely. Starting from $t=8.6$ (green zone), the user decides to grasp and moves the robot from pre-grasp to grasp. By comparing the region of the zone with entropy visualized (shared control mode) and the gray zone (auto mode), the design of the asynchronous shared control mechanism decreases the number of commands required from the user to fulfill the task, which is also aligned with the result shown in Fig.~\ref{fig: performance}.

\noindent \textbf{Analysis of Failure Cases.} Fig.~\ref{fig:failure_sub} illustrates a failure case. The text instruction is ``Grasp the pill bottle from the left''. At the beginning ($t=0.0$ to $4.3$), the user moves the robot towards the upper-right of the pill bottle. From $t=4.1$ to $9.9$, RT-V3 mispredicted that the user wanted to grasp the pill bottle from the top and approach the pill bottle with a top-down grasp pose. From $t=9.9$ to $21.6$, the user tried to slightly move away from the pill bottle and change the pose, but RT-V3 always dragged the robot back to the top-down grasp pose. From $t=21.6$ to $32.8$, the user decided to keep rotating towards the right-approaching pose, avoiding being dragged back to the top-down pose. In this case, RT-V3 mispredicted the user's intent, providing assistance that conflicted with the user's commands.

\section{Limitations and Future Work}
This paper introduced RT-V3, a probabilistic shared-control method for $SE(3)$ manipulation that utilizes a Bayesian framework to infer user intent through posterior estimation. We validated the approach across trajectory prediction, planning, and shared-autonomy tasks. Despite these strengths, several limitations remain to be addressed in future work.

Despite these strengths, several limitations remain:

\textbf{Distributional shift between behaviors of the path planner and the user.} Similar to its predecessor RT-V2, RT-V3 is trained to imitate the behavior of a path planner but is deployed as a prior distribution to estimate user behavior during posterior inference. This distributional shift between the automated planner and a human operator can lead to misalignments between RT-V3’s predicted actions and the user’s true intent. This limitation is also reflected in the failure cases shown in Fig.~\ref{fig:failure_sub}, where the system occasionally biases the motion toward an incorrect grasp hypothesis when the user's control strategy deviates from the planner's demonstrations. Future work could mitigate this issue in several ways. One promising direction is to incorporate human teleoperation data during training so that the learned prior better reflects real user behaviors rather than purely planner-generated demonstrations. Another direction is to introduce adaptive assistance mechanisms that monitor the consistency between the posterior intent estimate and the user’s control commands. For instance, when the posterior uncertainty becomes high or when the predicted assistance repeatedly conflicts with user inputs, the system could reduce the level of autonomous guidance or temporarily revert to direct user control. Such mechanisms could improve robustness to distributional mismatch while preserving the benefits of probabilistic intent inference.

\textbf{Reliance on the grasp planner.} RT-V3 is trained to approach a set of candidate grasp poses. Consequently, its performance is inherently bounded by the grasp planner’s accuracy. System efficacy is compromised if the planner (i) fails to generate the specific grasp poses intended by the user, or (ii) predicts the grasp poses that cannot make a successful pickup. However, we expect that this problem will be alleviated in the future as the grasp planners become stronger and stronger.

\textbf{Limited modeling capacity.} Learning a probabilistic reactive policy that reaches multiple candidate poses is a highly multi-modal task. Because shared control requires low-latency, real-time responses, the model size is strictly constrained. This limited capacity may prevent RT-V3 from capturing the full range of human multimodality, potentially degrading the user experience. Furthermore, assigning low probability density to action sequences that lead to collisions remains a challenge, as learning barrier functions within GMM distributions is difficult. As a result, collision-free behavior cannot be strictly guaranteed in the RT-V3 framework.

\textbf{Limited to grasping task.} RT-V3 is trained to approach the grasp, which means that it can only assist with the grasping task. To achieve general task assistance, the model must be trained on various tasks. Currently, vision-language-action models (VLAs) are developing rapidly and becoming increasingly accessible. We believe that the shared control mechanism proposed in the work combined with VLAs can inspire the community to develop a universal assistive controller.

\section{Conclusion}
In this work, we presented \emph{Robot Trajectron V3} (RT-V3), a Bayesian shared-control framework for $SE(3)$ manipulation that models user intent via a learned prior and computes a posterior over candidate actions to produce assistance. RT-V3 encodes point clouds and grasp poses predicted by the grasp planner as contextual input via a transformer-based architecture, and leverages a CVAE-GMM framework to model the multi-modality of the user's intended future actions. We propose to factorize the $\mathfrak{se}(3)$ action distribution into a translational distribution and a rotation distribution conditioned on translation, reducing the complexity of modeling user behavior.

Experimental results show that RT-V3 yields accurate trajectory predictions and competitive planning capability. In the shared autonomy experiments, RT-V3 provides smoother and faster assistance across different simulated users and human users, thereby improving success rates and reducing the cognitive burdens on users.

\appendix

\begin{figure*}[!t]
  \centering
  \includegraphics[width=0.7\linewidth]{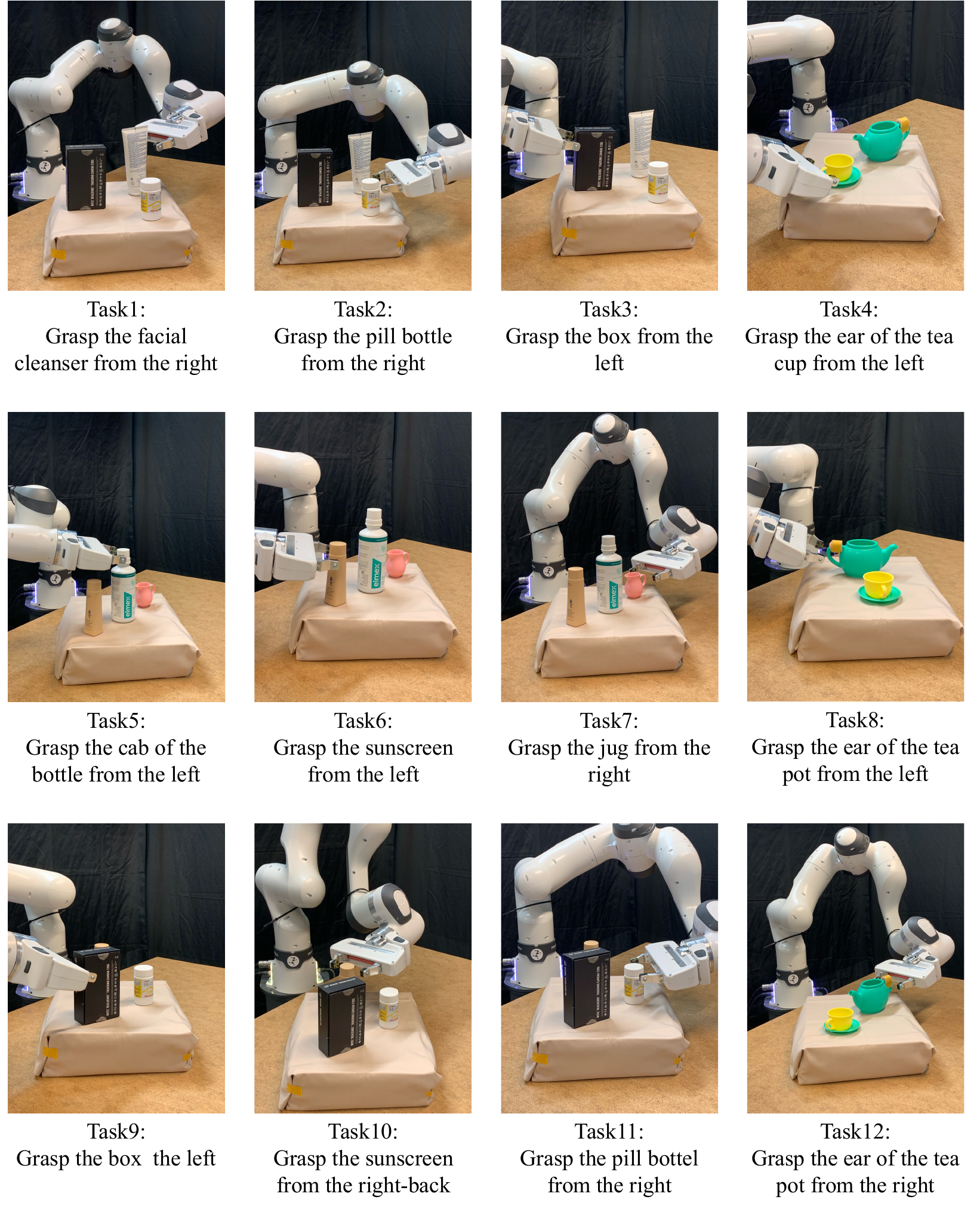}
  \caption{Twelve tasks in the real-world shared autonomy experiment. During the experiment, these tasks were separated into 3 groups (1-4, 5-8, 9-12), and three methods (Direct, HO, and RT-V3) were randomly assigned to one of the groups.}
  \label{fig: scenes}
\end{figure*}

\subsection{Proof for $SE(3)$ Equivariance of RT-V3}
We can abstract RT-V3 as a function $f$, which takes the relative past trajectory, point clouds, and grasp poses as input, as:
\begin{equation}
    \bm{a}^{\textnormal{rel}}_{0:H-1} = f(\bm{T}_{\text{cur}}^{\text{base}} \bm{x}_{-T+1:0}, \bm{T}_{\text{cur}}^{\text{base}} \mathcal{P}, \bm{T}_{\text{cur}}^{\text{base}} \mathcal{G}),
\end{equation}
where $\bm{T}_{\text{cur}}^{\text{base}} = (\bm{T}_{\text{base}}^{\text{cur}})^{-1}$ is the base frame expressed relative to the robot’s current end-effector frame, and the relative action sequence $\bm{a}^{\textnormal{rel}}_{0:H-1} = \bm{T}_{\text{cur}}^{\text{base}} \bm{a}_{0:H-1}$. Let us consider adding a random perturbation $\bm{T}_{\textnormal{rand}}$ to all the input data. Then the perturbed current end-effector frame becomes $\tilde{\bm{T}}_{\text{cur}}^{\text{base}} = (\bm{T}_{\text{rand}} \bm{T}^{\text{cur}}_{\text{base}})^{-1} =  \bm{T}_{\text{cur}}^{\text{base}} \bm{T}_{\text{rand}}^{-1}$, the perturbed past trajectory becomes $\tilde{\bm{x}}_{-T+1:0} = \bm{T}_{\text{rand}} \bm{x}_{-T+1:0}$, and the perturbed point clouds and grasp poses becomes $\tilde{\mathcal{P}} = \bm{T}_{\text{rand}} \mathcal{P}$ and $\tilde{\mathcal{G}} = \bm{T}_{\text{rand}} \mathcal{G}$, respectively. The RT-V3 prediction based on the perturbed data will be unchanged, as:
\begin{equation}
\begin{aligned}
    & f(\tilde{\bm{T}}_{\text{cur}}^{\text{base}} \tilde{\bm{x}}_{-T+1:0}, \tilde{\bm{T}}_{\text{cur}}^{\text{base}} \tilde{\mathcal{P}}, \tilde{\bm{T}}_{\text{cur}}^{\text{base}} \tilde{\mathcal{G}})\\
    & = f(\bm{T}_{\text{cur}}^{\text{base}} \bm{T}_{\text{rand}}^{-1} \bm{T}_{\text{rand}} \bm{x}_{-T+1:0}, \bm{T}_{\text{cur}}^{\text{base}} \bm{T}_{\text{rand}}^{-1} \bm{T}_{\text{rand}} \mathcal{P}, \bm{T}_{\text{cur}}^{\text{base}} \bm{T}_{\text{rand}}^{-1} \bm{T}_{\text{rand}} \mathcal{G}) \\
    & = f(\bm{T}_{\text{cur}}^{\text{base}} \bm{x}_{-T+1:0}, \bm{T}_{\text{cur}}^{\text{base}} \mathcal{P}, \bm{T}_{\text{cur}}^{\text{base}} \mathcal{G}) \\
    & = \bm{a}^{\textnormal{rel}}_{0:H-1}
\end{aligned} 
\end{equation}
Through transforming $\bm{a}^{\textnormal{rel}}_{0:H-1}$ by the perturbed current end-effector's frame, we obtain the perturbed action $\bm{a}_{0:H-1}= \tilde{\bm{T}}_{\text{cur}}^{\text{base}}  \bm{a}^{\textnormal{rel}}_{0:H-1}$. Therefore, we prove the $SE(3)$ equivariance of RT-V3.

\subsection{Distance Metric}
Given two $\textnormal{SE}(3)$ poses $T_A$ and $T_B$, we can transform them into translation positions $p_A$, $p_B$ and Euler angles $a_A$ and $a_B$. We define the manifold distance between two poses as:
\begin{equation}
    d(T_A, T_B) =  \left\|  p_A-p_B\right\|_1 + \left\| \frac{0.3}{\pi}*(a_A-a_B) \right\|_1.
\end{equation}
Considering the size of the workspace is $0.3 \times 0.3 \times 0.3$ m$^3$, the scaling term $\frac{0.3}{\pi}$ balances the translation and rotation errors.

\begin{figure*}[t!]
\centering
    \setlength{\belowcaptionskip}{-10pt} 
    
    \subfloat[\label{fig:scene1_3}]{
        \includegraphics[width=0.9\textwidth]{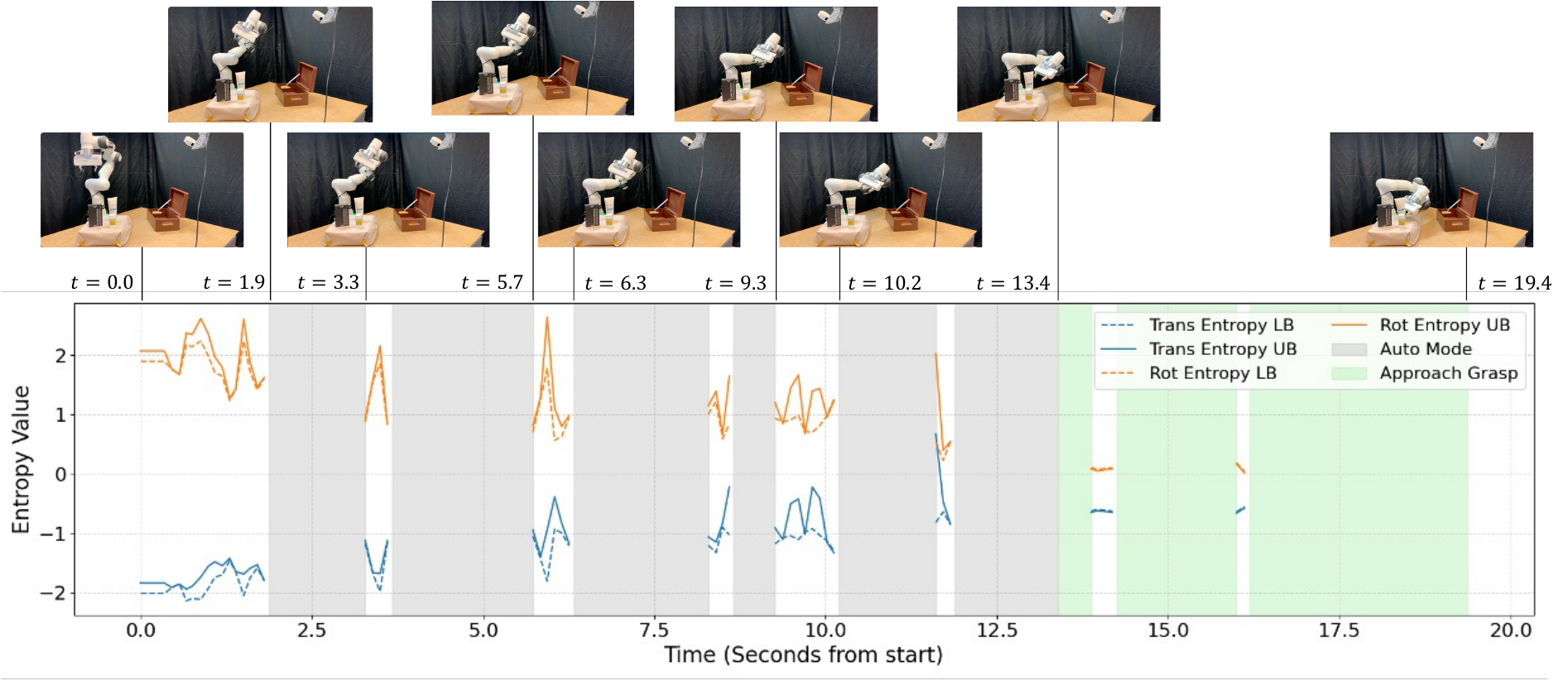} 
    } \\ 
    \vspace{-1em} 
    
    \subfloat[\label{fig:scene1_1}]{
        \includegraphics[width=0.9\textwidth]{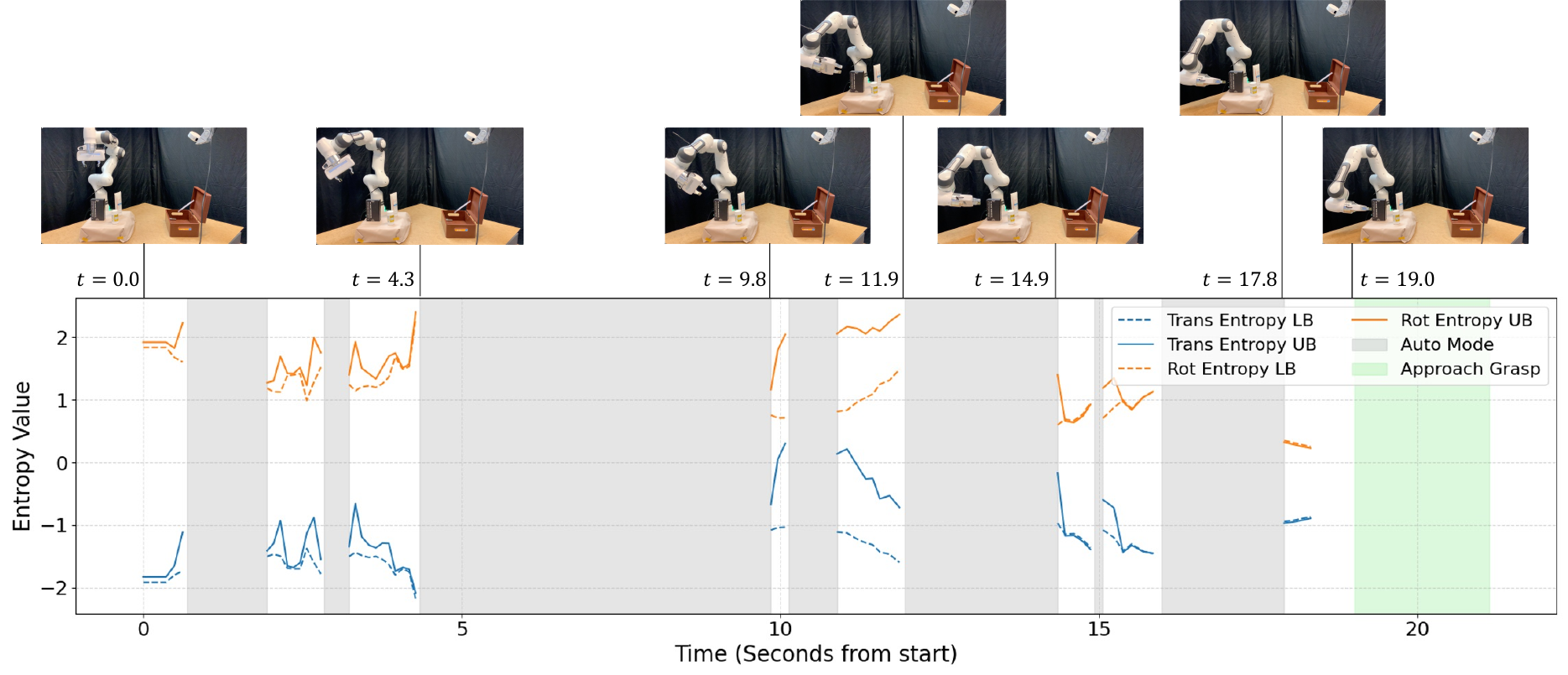}
    } \\
    \vspace{-0.8em}
    
    
    \subfloat[\label{fig:scene2_2}]{
        \includegraphics[width=0.9\textwidth]{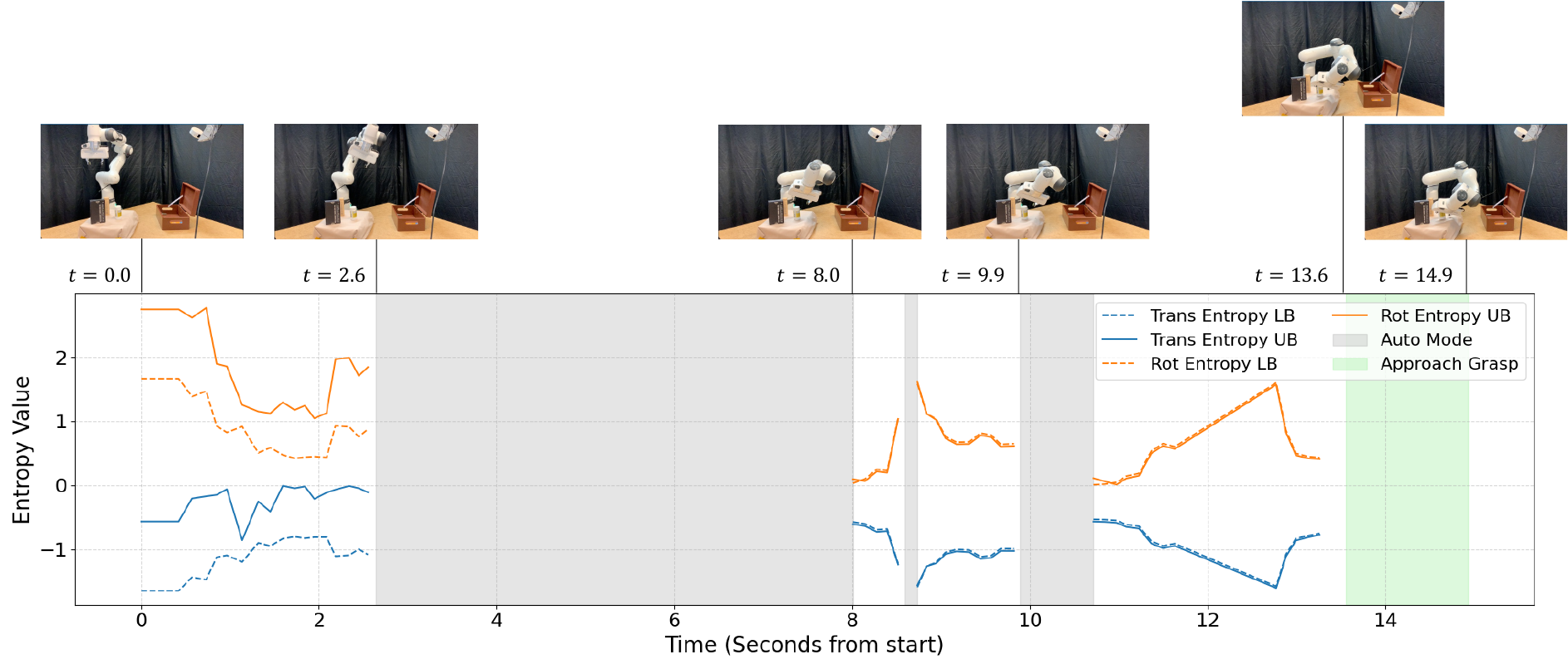}
    }
    
    \caption{Visualizations of four different demonstrations. (a) Task1: Grasp the facial cleanser from the right. (b) Task3: Grasp the box from the left. (c) Task11: Grasp the pill bottle from the right.
    }
\label{fig: more demos}
\end{figure*}


\subsection{Simulated User for Shared Control}
To rigorously evaluate RT-V3's shared control capabilities, we implemented five distinct simulated users. These users represent common limitations found in real-world human-machine interfaces, such as noise, latency, and restricted degrees of freedom (DoF). All users generate commands in the form of a 6D twist vector $\mathbf{u} = [v_x, v_y, v_z, \omega_x, \omega_y, \omega_z]^\top \in \mathfrak{se}(3)$, representing the velocity required to move from the current end-effector pose $\mathbf{T}_{\textnormal{cur}}$ to a goal pose $\mathbf{T}_{\textnormal{goal}}$ in the local frame, as:
\begin{equation}
    \mathbf{u} = \textnormal{Logmap}(\mathbf{T}_{\textnormal{goal}} \mathbf{T}_{\textnormal{cur}}^{-1}).
\end{equation}

\subsubsection{Noisy User}
The Noisy User simulates a human operator with imprecise motor control or a high-variance input device. The command is generated by adding Gaussian noise to the ideal twist:
\begin{equation}
    \mathbf{u} = \mathbf{e} + \epsilon, \quad \epsilon \sim \mathcal{N}(0, \Sigma),
\end{equation}
where $\Sigma$ is a diagonal covariance matrix. This tests the model's ability to filter high-frequency noise and extract the underlying intent.

\subsubsection{Laggy User}
The Laggy User simulates network latency or delayed human reaction times. At each timestep, the user has an 80\% probability of repeating the previous command $\mathbf{u}_{t-1}$ rather than calculating a new error-correcting twist. This creates ``sticky'' behavior where the input does not immediately update as the robot moves.

\subsubsection{Mode-Switching User}
The Mode-Switching User alternates between purely translational and purely rotational control. (i) Translation Mode: $\mathbf{u} = [v_x, v_y, v_z, 0, 0, 0]^\top$. (ii) Rotation Mode: $\mathbf{u} = [0, 0, 0, \omega_x, \omega_y, \omega_z]^\top$. The user switches modes every 20 steps (2 seconds in simulation). This evaluates if the model can maintain the ``passive'' components of the pose while the user focuses on the ``active'' components.

\subsubsection{Single-DoF User}
The Single-DoF User represents the most constrained input scenario (e.g., a single-axis joystick or button control). Every 10 steps (1 second in simulation), the user identifies the single DOF with the largest absolute error:
\begin{equation}
    i^* = \arg\max_i | \mathbf{e}_i |
\end{equation}
For the duration of that period, the user only provides input for that specific index:
\begin{equation}
    \mathbf{u}_i = \begin{cases} \mathbf{e}_i & \text{if } i = i^* \\ 0 & \text{otherwise}, \end{cases}
\end{equation}
This requires the shared control system to provide significant assistance in the remaining 5 DOFs to ensure smooth motion.

\subsection{Twelve tasks in the real-world shared autonomy experiment}
Fig.~\ref{fig: scenes} shows 12 tasks in the real-world shared autonomy experiment. These tasks include grasping different objects or different parts of the object from different orientations.

\subsection{More visualizations of Guidance in Real-world Shared Autonomy Experiments}
To further demonstrate the robustness and adaptability of RT-V3 across diverse manipulation scenarios, we provide additional visualizations of entropy and task execution for four distinct tasks in Fig.~\ref{fig: more demos}. Across these four tasks, ``auto mode'' occupies a large time area, which denotes, similar to what we discuss in the Sec.~\ref{sec: real-world exp}, that RT-V3 decreases the user commands by accurately predicting and automatically executing the intended actions. In addition, these demonstrations showcase how the Bayesian framework manages multi-modal intent under varying degrees of task complexity.
\begin{itemize}
\item \textbf{A increasing trend of translational entropy:} We observe that as the task proceeds, the translational entropy will have an increasing trend. This indicates that at the beginning of the trial, the reaching motion is relatively certain since the robot has to first approach the objects. Once the robot gets close to the object, many candidate grasp poses are densely distributed on the surface of the object, indicating that the robot could reach any surrounding grasp poses, leading to a high entropy.

\item \textbf{A decreasing trend of translational entropy:} In contrast to the translational entropy, the rotational entropy will have a decreasing trend as the task proceeds. That is because when the robot gets close to the objects, many of the surrounding grasp poses, which are predicted by the grasp planner, have similar orientations. Thus, the entropy is relatively low.

\item \textbf{Multi-modal action prediction:} In Fig.~\ref{fig:scene2_2}, the difference between the entropy upper bound and lower bound is relatively large at the beginning of the trial. This implies that RT-V3 is uncertain about the user's intent and predicts a highly multi-modal action distribution. Thus, we can also observe that the entropy values reach a high level. (Trans Entropy UB is higher than 4, and Rot Entropy UB is higher than 3). 
\end{itemize}


\bibliographystyle{IEEEtran}
\bibliography{IEEEexample}

\end{document}